%% file: _main.tex
\input{_constants}
\arxiv 

\pdfoutput=1
\documentclass[10pt,twocolumn,letterpaper]{article}
\input{cvpr_header}

\unless\ifarxiv \myexternaldocument{_supplementary} \fi

\begin{document}
\title{\paperTitle}
\author{\authorBlock}
\maketitle

\input{00_abstract}

\input{01_intro}

\input{02_related}

\input{03_method}
\input{04_exp}

\input{10_conclusion}

{\small
\bibliographystyle{ieeenat_fullname}
\bibliography{11_references}
}

\ifarxiv \clearpage \appendix \input{12_appendix} \fi

\end{document}

%% file: _constants.tex
\def\paperID{}
\def\confName{}
\def\confYear{}

\def\paperTitle{\textit{Exact}: Exploring Space-Time Perceptive Clues for Weakly Supervised\\ Satellite Image Time Series Semantic Segmentation}

\def\authorBlock{
    
    Hao Zhu$^1$, 
    Yan Zhu$^1$, 
    Jiayu Xiao$^1$, 
    Tianxiang Xiao$^1$, 
    Yike Ma$^1$, 
    Yucheng Zhang$^1$, 
    Feng Dai$^1$\\
    $^1$Institute of Computing Technology, Chinese Academy of Sciences, Beijing, China \\
    {\tt\small zhuhao22z@ict.ac.cn}
}

\newif\ifreview 
\newif\ifarxiv \newcommand{\arxiv}{\arxivtrue}
\newif\ifcamera 
\newif\ifrebuttal 

%% file: cvpr_header.tex
\ifreview \usepackage[review]{cvpr} \fi
\ifarxiv \usepackage[pagenumbers]{cvpr} \fi
\ifrebuttal \usepackage[rebuttal]{cvpr} \fi
\ifcamera \usepackage{cvpr} \fi

\input{_macros}  

\usepackage{xr-hyper}

\makeatletter
\newcommand*{\addFileDependency}[1]{
  \typeout{(#1)}
  \@addtofilelist{#1}
  \IfFileExists{#1}{}{\typeout{No file #1.}}
}

\makeatother
\newcommand*{\myexternaldocument}[1]{
    \externaldocument{#1}
    \addFileDependency{#1.tex}
    \addFileDependency{#1.aux}
}

\definecolor{cvprblue}{rgb}{0.21,0.49,0.74}
\usepackage[pagebackref,breaklinks,colorlinks,allcolors=cvprblue]{hyperref}
\usepackage[capitalize]{cleveref}
\crefname{section}{Sec.}{Secs.}
\crefname{table}{Table}{Tables}
\crefname{figure}{Fig.}{Figs.}

\ifarxiv \crefname{appendix}{App.}{Apps.}
\else \crefname{appendix}{Suppl.}{Suppls.} \fi

%% file: _macros.tex
\usepackage{graphicx}	
\usepackage{amsmath}	
\usepackage{amssymb}	
\usepackage{booktabs}
\usepackage{times}
\usepackage{microtype}
\usepackage{epsfig}
\usepackage{caption}
\usepackage{float}
\usepackage{placeins}
\usepackage{color, colortbl}
\usepackage{stfloats}
\usepackage{enumitem}
\usepackage{tabularx}
\usepackage{xstring}
\usepackage{multirow}
\usepackage{xspace}
\usepackage{url}
\usepackage{subcaption}
\usepackage{xcolor}
\usepackage[hang,flushmargin]{footmisc}
\usepackage{pifont}
\usepackage{bbding}
\usepackage{bbm}
\usepackage{mathtools}
\usepackage{algorithm}
\usepackage{algorithmic}

\ifcamera \usepackage[accsupp]{axessibility} \fi

% ...

\ifarxiv  \fi

\newcommand{\R}[1]{{%
    \textbf{%
        \ifstrequal{#1}{1}{\textcolor{red}{R#1}}{%
        \ifstrequal{#1}{2}{\textcolor{blue}{R#1}}{%
        \ifstrequal{#1}{3}{\textcolor{magenta}{R#1}}{%
        \ifstrequal{#1}{4}{\textcolor{teal}{R#1}}{%
                           \textcolor{cyan}{R#1}%
        }}}}%
    }%
}}

%% file: 00_abstract.tex
\begin{abstract}

Automated crop mapping through Satellite Image Time Series (SITS) has emerged as a crucial avenue for agricultural monitoring and management.
However, due to the low resolution and unclear parcel boundaries, annotating pixel-level masks is exceptionally complex and time-consuming in SITS. 
This paper embraces the weakly supervised paradigm (i.e., only image-level categories available) to liberate the crop mapping task from the exhaustive annotation burden.
The unique characteristics of SITS give rise to several challenges in weakly supervised learning: (1) noise perturbation from spatially neighboring regions, and (2) erroneous semantic bias from anomalous temporal periods.
To address the above difficulties, we propose a novel method, termed \textbf{ex}ploring sp\textbf{ac}e-\textbf{t}ime perceptive clues (\textbf{Exact}).
First, we introduce a set of spatial clues to explicitly capture the representative patterns of different crops from the most class-relative regions. 
Besides, we leverage the temporal-to-class interaction of the model to emphasize the contributions of pivotal clips, thereby enhancing the model perception for crop regions. 
Build upon the space-time perceptive clues, we derive the clue-based CAMs to effectively supervise the SITS segmentation network.
Our method demonstrates impressive performance on various SITS benchmarks. 
Remarkably, the segmentation network trained on  Exact-generated masks achieves \textbf{95\%} of its fully supervised performance, showing the bright promise of weakly supervised paradigm in crop mapping scenario.  
Our code will be publicly available \href{https://github.com/MiSsU-HH/Exact}{here}.
\let\thefootnote\relax\footnote{\scriptsize{Preprint. Under review.}}

\end{abstract}

%% file: 01_intro.tex
\section{Introduction}
\label{sec:intro}

\begin{figure}[t]
\centering
\includegraphics[width=1.0\linewidth]{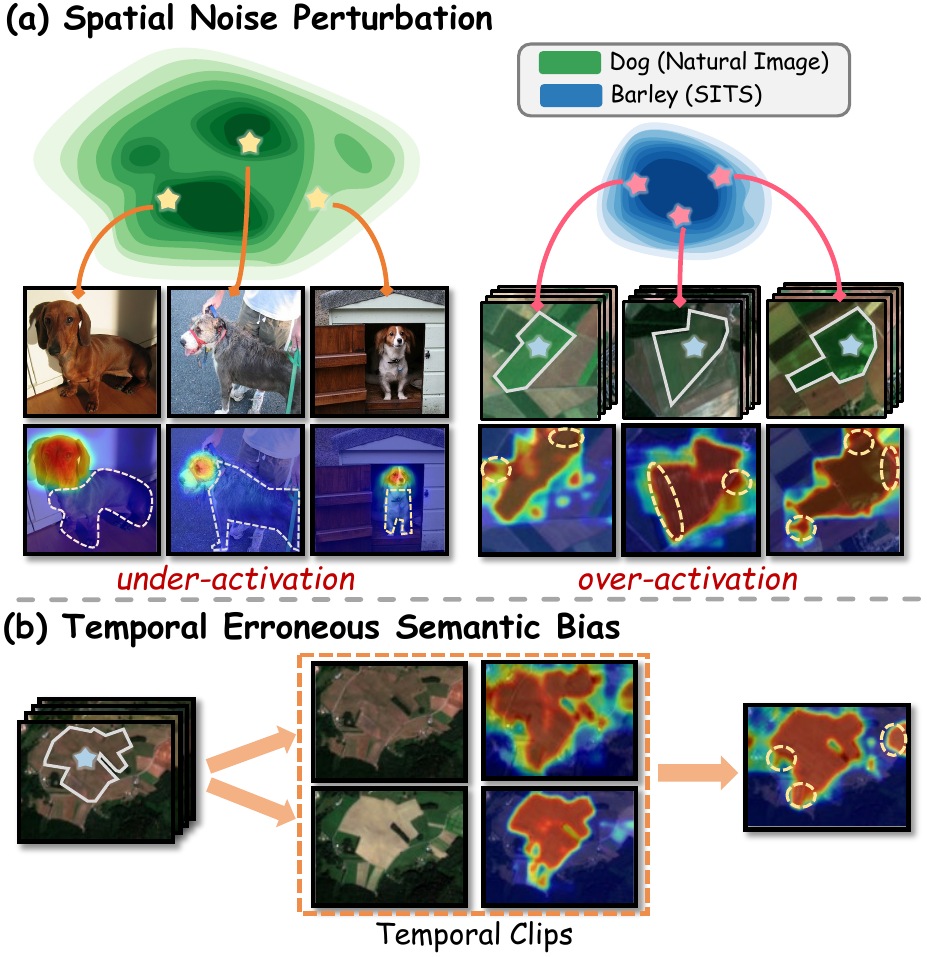}
\caption{
\textbf{Illustration of the two inherent issues arised from spatial and temporal perspectives in SITS.} (a) shows noise pertubation from the spatial perspective. We visual the high-level feature manifold of \textcolor[RGB]{57,162,86}{Dog} (natural image) and \textcolor[RGB]{43,123,186}{Barley} (SITS) to reveal the distinct spatial properties. The feature dimensions are reduced by t-SNE~\cite{tsne}.
(b) shows the erroneous semantic bias induced by anomalous temporal clips. We denote the parcel regions with $\bigstar$.
}
\label{fig:intro}
\end{figure}

The launch of numerous public and commercial satellites provides broader opportunities to record, analyze, and predict the evolution of crop land~\cite{s1,s2,landsat,omisat}. 
In this context, Satellite Image Time Series (SITS) with $10\texttt{m}$ resolution offered by the high-frequency Sentinel-2 (S2) satellites serve as a valuable data source for automated crop mapping~\cite{cropmapping,cropsupervision,forecast,skysense,satmae,s2mae}. 
The core of crop mapping lies in the semantic segmentation of crop parcels. 
Recently, many efforts are devoted to exploiting the versatile and powerful relation modeling capability of deep neural networks for this task~\cite{tae,pastis,tsvit,fpn_clstm}. 
While such methods have shown a significant progress, they rely heavily on pixel-level manual annotation, which is notoriously complex and time-consuming~\cite{weakrs}. The low resolution of satellite images and the indistinct boundaries between crop parcels complicate the annotation process. Even worse, the varying phenological cycles among crops require annotators to meticulously select appropriate acquisition times for annotation.

To address this, one promising solution is weakly supervised semantic segmentation (WSSS), which relies solely on time-efficient annotation form, \ie, image-level categories.
The image-level WSSS has commonly studied in natural image domain. 
Existing methods primarily extract Class Activation Map (CAM)~\cite{cam} to generate pseudo labels for training semantic segmentation models.
The intention of CAM is to identify the regions that highly contribute to the prediction of each class, revealing the shared patterns among the images within the same category. In context of natural images, CAM tends to highlight local discriminative parts of the target object. The reason is that the dispersed intra-class distribution enforces the model to learn a extremely sharp decision boundary, thereby the classifier weights tend to interact with feature representations within more discriminative object regions, as shown in \cref{fig:intro}\textcolor[RGB]{42,120,189}{a} left. Thus the primary focus of researchers falls into expanding the CAM to identify 
the entire object semantics for natural images.

Different from WSSS for natural image, crop mapping for SITS mainly faces with two challenges: (1) from the spatial perspective, parcel objects within the same category share uniform appearance and color, exhibiting strong neighboring consistency. The intra-class compactness of local patterns looses the tolerance of the classifier to noise perturbation, resulting in pronounced over-activation phenomena in CAM, as shown in \cref{fig:intro}\textcolor[RGB]{42,120,189}{a} right.
The disparate characteristic lead to the advances in natural image domain cannot directly benefit the SITS crop segmentation. 
(2) From the temporal perspective, although different crops show distinct phenological cycles and varying characteristics, 
they may display similar appearance in some specific periods. This confusion imbues wrong semantic bias to the learning process of the model, thereby activating some undesired semantic regions in CAM.
An intuitive illustration is shown in \cref{fig:intro}\textcolor[RGB]{42,120,189}{b}, the temporal clips that deviate from the pivotal semantic affect the perceptual ability of CAM to correct crop regions. 

In this work, we present \textit{\textbf{ex}ploring sp\textbf{ac}e-\textbf{t}ime perceptive clues (\textbf{Exact})}, a tailored WSSS framework for crop mapping, to cope with the challenges arised from the spatial and temporal aspects respectively. Firstly, we introduce a set of spatial clues to explicitly capture the patterns of different crops. Leveraging the filtered CAM as an indicator, we update representative clues by conducting spatial clustering from the most class-relative regions.
These clues are then used to regularize the feature space via optimizing the contrastive objective, thereby sharpening the decision boundary of the model and mitigating the perturbations from illusory patterns.
Secondly, to cope with the erroneous semantic bias caused by anomalous temporal periods, we propose temporal-aware affinity propagation to emphasize the contributions of pivotal clips for crop perception.
In detail, we extract the temporal-to-class attention from the model to reweight the temporal sequence embeddings. The modulated representations can be used to model temporal-aware pairwise affinity for propagation on the raw CAM, thus effectively suppressing the undesired semantic regions in a self-supervised manner.

Unlike existing WSSS methods that rely on classifier weights to generate CAMs, we exploit the well-updated space-time perceptive clues to derive the clue-based CAMs (CB-CAMs) as pseudo labels for segmentation. 
Compared to the raw CAM, the CB-CAMs (1) remarkably suppress the perturbations from spatial and temporal aspects, and (2) delineate the crop regions more precisely, thereby providing more reliable supervision for the subsequent segmentation.

Our main contributions can be summarized as follows:
\begin{itemize}
	\item  
	We introduce the WSSS paradigm to SITS crop mapping task to tackle the daunting annotation challenge. 
	To the best of our knowledge, this is the first work that relies solely on image-level categories for crop segmentation. 
	\vspace{2pt}
	\item 
    To overcome the drawbacks arised from the spatial and temporal aspects of SITS, we propose \textit{\textbf{Exact}} that explores space-time perceptive clues to reduce the noise perturbation and rectify the wrong semantic bias, ultimately providing reliable supervision for SITS segmentation.
	\vspace{2pt}
	\item We experimentally show that \textit{Exact} achieves impressive performance on common benchmarks. Using \textit{Exact}-generated labels for training, SITS segmentation model attains up to \textbf{95\%} of its fully supervised performance. Our results significantly advance the upper bound of image-level WSSS technique compared to the other domains. 
\end{itemize}

%% file: 02_related.tex
\section{Related Work}
\label{sec:related}

\noindent\textbf{Semantic segmentation on SITS.}~Automated crop monitoring through Satellite Image Time Series (SITS) has attracted great interest among researchers and demonstrated considerable social impact~\cite{sits1,sits2,sits3,sits4,skysense,satmae}. 
One of the challenging tasks is the semantic segmentation of agricultural parcels. The goal of the network is to learn a mapping function that assigns each pixel in the SITS to the corresponding crop type or background. 
Some works process SITS inputs by first extracting spatial information and then compressing the temporal dimension.
For example, U-ConvLSTM~\cite{unet2d} relied on U-Net~\cite{unet} architecture to encode the spatial dimension, followed by a ConvLSTM~\cite{convlstm} for the temporal dimension. Similarly, the FPN-ConvLSTM~\cite{fpn_clstm} replaced the U-Net with a Feature Pyramid Network~\cite{fpn} as the spatial encoder. U-TAE~\cite{pastis} compressed the temporal dimension through the outstanding temporal attention mechanism~\cite{tae}.
The other option to encode SITS is the \textit{temporal-spatio} scheme, which first processes the temporal dimension and then extracts the spatial information.
Ru{\ss}wurm \etal~\cite{germany} employed bidirectional LSTM to extract temporal features and then used CNNs to integrate spatial information. 
Recently, TSViT proposed to borrow the powerful dependency modeling capability of the Vision Transformer (ViT)~\cite{vit} to handle SITS, achieving state-of-the-art performance at lower computational cost. The TSViT also comprehensively illustrated the superiority of the temporal-spatio scheme.
Taking a holistic view, we adopt this scheme to process SITS, aiming to achieve an optimal trade-off for automated crop mapping.

\begin{figure*}[t]
\centering
\includegraphics[width=1.0\linewidth]{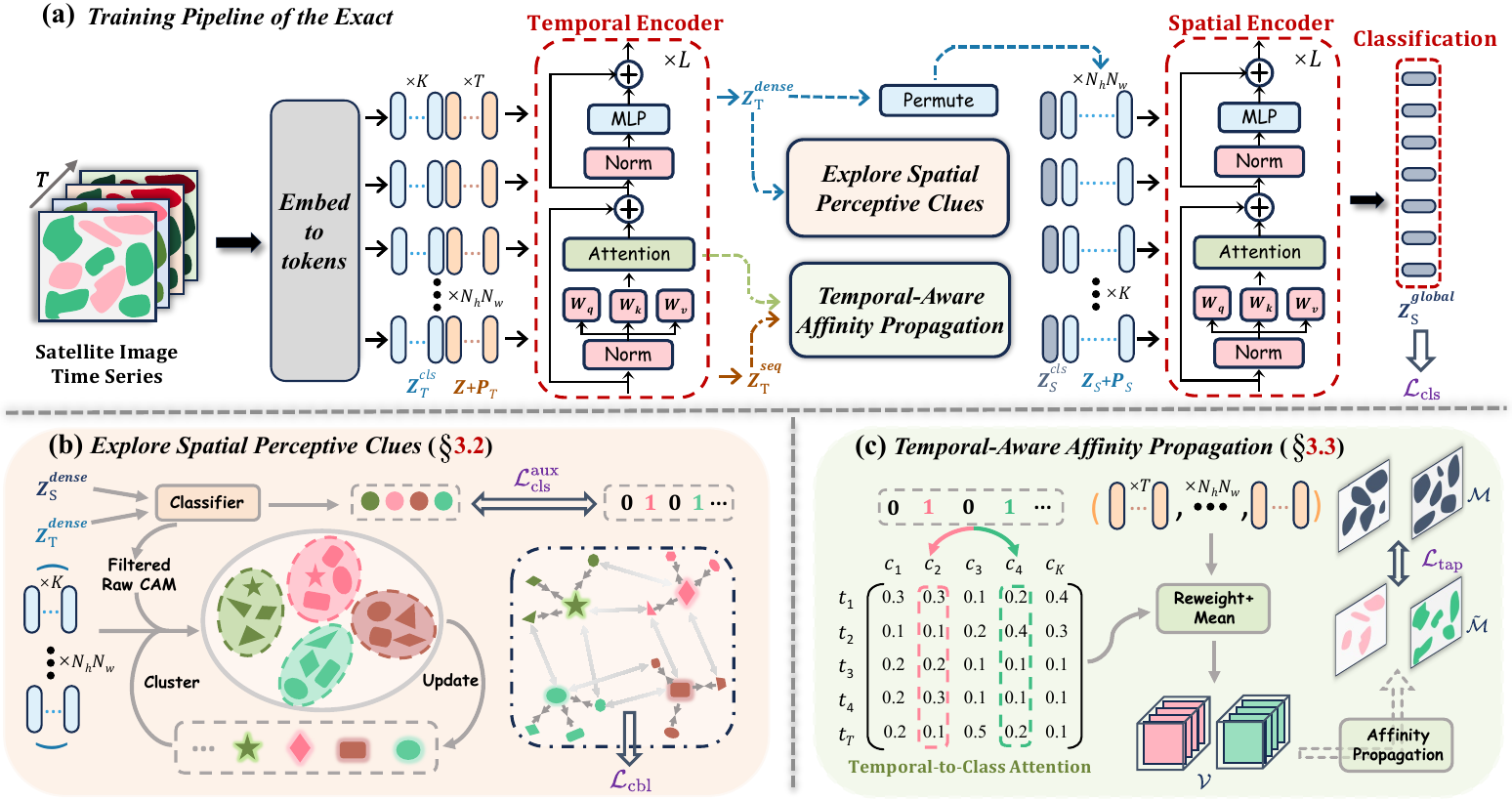}
\caption{(a) \textbf{The training pipeline of \textit{Exact}}. We adopt the Temporal-Spatio scheme to handle the SITS input, which contains two transformer encoders. The first temporal encoder models interactions between acquisition times, then the followed spatial encoder discards the temporal dimension and models interactions between spatial positions. To overcome the difficulties arised from spatial and temporal aspects, we propose two novel technologies in temporal embedding space: (b) \textbf{Explore Spatial Perceptive Clues} to mitigate the noise perturbation (see $\S$\ref{sec:clue}) and (c) \textbf{Temporal-Aware Affinity Propagation} to rectify the wrong semantic bias(see $\S$\ref{sec:affinity}).  }
\label{fig:overview}
\end{figure*}

\noindent\textbf{Weakly supervised semantic segmentation.}~ Weakly supervised semantic segmentation (WSSS) with image-level labels (\ie ground-truth object categories) has shown significant success in natural images~\cite{lpcam,frozen_clip,ktse,mctformer,cti,mctformer+}.
Most advanced WSSS methods follow the three-steps pipeline of: 1) training a classification network with image-level labels, 2) obtaining class activation map (CAM)~\cite{cam,grad_cam,grad_cam++} from the well-optimized classification network as pixel-level coarse labels, 3) refining the coarse labels to train the final semantic segmentation network. 
Since the CAM generation is the foundation step of the entire pipeline, numerous works has been proposed to address the under- and over-activation issues of the initial CAM~\cite{seam,seco,ppc,sipe,cpal,toco,fpr}. 
To derive the final pseudo labels, CAM requires cumbersome post-processing, including random walks~\cite{psa,irn} and denseCRF~\cite{crf} refinement. 
Thanks to the excellent works, the performance of image-level WSSS in natural images currently achieves 90\% of pixel-level supervision.
However, the WSSS networks designed for natural images require significant adaptation to be applied to SITS, and the results are still unsatisfactory due to the distinct data characteristics. In this work, we incorporate WSSS technique into the SITS and overcome the difficulties caused by inherent data properties.

%% file: 03_method.tex
\section{Method}
In this section, we first look more closely at the temporal-spatio scheme and class activation map technique ($\S$\ref{sec:revist}). Next, we introduce how to explore spatial perceptive clues ($\S$\ref{sec:clue}) and conduct temporal-aware affinity propagation ($\S$\ref{sec:affinity}). Finally, we show the overall objective function and the clue-based CAM generation strategy of \textit{Exact} ($\S$\ref{sec:overall}). 
The training pipeline of \textit{Exact} is shown in \Cref{fig:overview}. 
\label{sec:method}
\subsection{Preliminaries}
\label{sec:revist}
\noindent\textbf{The temporal-spatio scheme} that first processing the temporal dimension and then extracting the spatial information has shown significant superiority when dealing with SITS data. Following the TSViT~\cite{tsvit}, we use the variant ViT~\cite{vit} as the backbone for both temporal and spatial encoders. We consider a SITS input $\mathbf{X}\in\mathbb{R}^{T\times C\times H\times W}$ with $T$ the length of time series, $C$ the number of channels and $H\times W$ the spatial dimensions. The input $\mathbf{X}$ is mapped into a sequence of patch tokens and reshaped to $\mathbf{Z}\in\mathbb{R}^{ N_h\cdot N_w \times T\times d}$, where $N_h=[\frac{H}{h}]$, $N_w=[\frac{W}{w}]$ and $h\times w$ is the spatial extent for each patch. We then add the temporal position embeddings $\mathbf{P}_T\in \mathbb{R}^{T\times d}$ and concatenate the temporal multi-class tokens $\mathbf{Z}_T^\text{cls}\in \mathbb{R}^{K\times d}$ to obtain the input of temporal encoder:
\begin{align}
\mathbf{Z}_T^\text{in}=[\mathbf{Z}_{T}^\text{cls}, \mathbf{Z}+\mathbf{P}_T], \mathbf{Z}_T^\text{in}\in\mathbb{R}^{N_h\cdot N_w \times (K+T)\times d}
\end{align}
here $K$ denotes the number of categories, $\mathbf{Z}_T^\text{cls}$ and $\mathbf{P}_T$ are repeated $N_h\cdot N_w$ times to match the spatial shape.
With the output feature maps $\mathbf{Z}^\text{out}_T=[\mathbf{Z}^\text{dense}_T|\mathbf{Z}^\text{seq}_T]$ from the temporal encoder, we extract the first $K$ tokens $\mathbf{Z}^\text{dense}_T \in \mathbb{R}^{N_h\cdot N_w\times K \times d}$ and permute the first and second dimensions to serve as input patch tokens $\mathbf{Z}_S\in\mathbb{R}^{K \times N_h\cdot N_w\times d}$ for spatial encoder. 
These inputs are then combined with spatial multi-class tokens $\mathbf{Z}_{S}^\text{cls}\in\mathbb{R}^{K\times 1\times d}$ and spatial position embeddings $\mathbf{P}_S\in\mathbb{R}^{N_h\cdot N_w\times d}$ at all $K$ spatial representations: 
\begin{equation}
\mathbf{Z}_S^\text{in}=[\mathbf{Z}_{S}^\text{cls}, \mathbf{Z}_S+\mathbf{P}_S], \mathbf{Z}_S^\text{in}\in\mathbb{R}^{K \times (1+N_h\cdot N_w)\times d}
\end{equation}
After the spatial encoder phase, we separate the output features $\mathbf{Z}_S^\text{out}=[\mathbf{Z}^\text{global}_S|\mathbf{Z}^\text{dense}_S]$ to align with different downstream tasks. For classification task, we feed the global tokens $\mathbf{Z}_S^\text{global}\in\mathbb{R}^{K\times1\times d}$ into the classifier to obtain classification logits. For segmentation task, to derive the dense prediction mask, the dense tokens $\mathbf{Z}_S^\text{dense}\in\mathbb{R}^{K\times N_h\cdot N_w\times d}$ are fed into the segmentation decoder. 

\noindent\textbf{Class activation map} (CAM)~\cite{cam} is widely used in WSSS to provide weak annotations that rely on image-level labels. Given a natural image, its feature maps $\mathbf{F}\in\mathbb{R}^{H^{\prime}\times W^{\prime}\times D}$ are extracted by a classification backbone. To derive the classification score, the feature maps are average pooled and multiplied by the classifier weights $\boldsymbol{w} \in\mathbb{R}^{K\times D}$. CAM is generated by weighting and summing each channel in the feature maps with the classifier weights, as follows:
\begin{equation}
    \mathcal{M}^{k}_{i}=\mathtt{ReLU}\left(\sum_i\boldsymbol{w}^{k}_{i} \cdot \mathbf{F}_{:,:,i}\right),\quad\forall k\in K.
\end{equation}
Most WSSS methods normalize $\mathcal{M}^{k}$ to the range [0,1] and apply a global threshold to filter out background pixels to obtain finally pseudo labels. In the temporal-spatio network, we feed the dense tokens $\mathbf{Z}^\text{dense}_T$ and $\mathbf{Z}^\text{dense}_S$ into the classifier and compute the additional classification loss $\mathcal{L}_\text{cls}^\text{aux}$ to generate the fused raw CAM for SITS. 

\subsection{Explore Spatial Perceptive Clues}
\label{sec:clue}
\textbf{CAM filtering.} 
Given an input SITS $\mathbf{X}$ and its image-level label $y\in[0,1]^K$, we first compute its normalized fused CAM $\mathcal{M}\in\mathbb{R}^{N_h\cdot N_w\times K}$ by the classifier weights and the output dense tokens of temporal and spatial encoder. We use two threshold scores $\mu_l$ and $\mu_h$ to filter out the reliable foreground, background and uncertatin regions:
\begin{equation}
\hat{\mathcal{M}}=\begin{cases}0,&\mathrm{if} \ \mathcal{M}\leq\mu_{l},
\\1,&\mathrm{if} \ \mathcal{M}\geq\mu_{h},
\\\mathrm{ignore},&\mathrm{otherwise}. \end{cases}
\end{equation}
\textbf{Clues clustering.} 
To capture the compact patterns of different crops, we establish a group of class-wise representative prototypes, which later serve as perceptive clues to generate high-quality pseudo labels. 
In crop segmentation task, temporal features often provide more information than spatial context~\cite{tsvit}, so we choose to perform spatial clustering on temporal dense embeddings $\mathbf{Z}^\text{dense}_T$ over the whole dataset. 
Specifically, we build the class-wise positive prototype set $\mathcal{P}_\text{pos}=\{\boldsymbol{p}_\text{pos}^k\in\mathbb{R}^{N_p\times d}\}_{k=1}^K$ and negative prototype set $\mathcal{P}_\text{neg}=\{\boldsymbol{p}_\text{neg}^k\in\mathbb{R}^{N_p\times d}\}_{k=1}^K$, where $N_p$ denotes the number of prototypes. 
If $k$-th class appears in a training batch, we update the prototypes $\boldsymbol{p}^k$ via solving the optimal transport problem~\cite{swav,protoseg}. Given a mapping matrix $\mathbf{C}^k$ that represents the assignment between each pixel and its prototype, it can be referred as an element of the transportation polytope~\cite{self-label}:
\begin{equation}
\mathbb{C}=\{\mathbf{C}^k\in \mathbb{R}^{N_p\times N_k}_+|\mathbf{C}^k\mathbf{1}=\boldsymbol{u}, \mathbf{C}^{k\top}\mathbf{1}=\boldsymbol{r}\},
\end{equation}
where $N_k$ denotes the number of pixels belonging to class $k$. $\mathbf{1}$ represents the vectors of ones in appropriate dimensions, $\boldsymbol{u}$ and $\boldsymbol{r}$ are the marginal projections onto the rows and columns of $\mathbf{C}^k$.
We can maximize the following objective function to optimize the mapping matrix:
\begin{equation}
	\label{eq:6}
	{\small \begin{split}
			\underset{\mathbf{C}^{k}\in \mathbb{C}}{\rm maximize}\ {\mathtt{Tr}}(\mathbf{C}^{k\top}\boldsymbol{p}^k\mathbf{Z}^{k\top})-\eta\sum_{\mathclap{n_p,n_k\in N_p,N_k}}\mathbf{C}^k_{n_pn_k}\log\mathbf{C}^k_{n_pn_k},
	\end{split}}
\end{equation} 
here $\mathbf{Z}^k\in\mathbb{R}^{N_k\times d}$ is the corresponding temporal dense embeddings belonging to class $k$, $\eta$ controls the smoothness of entropy regularization term.  
The continuous approximate solution of Eq.~(\ref{eq:6}) can be obtained through iteratively applying the Sinkhorn-Knopp algorithm~\cite{sink}. Subsequently, we momentum update the $n_p$-th prototype of class $k$ according to the assignment matrix and the embeddings: 
\begin{equation}
\boldsymbol{p}_{n_p}^k=\alpha \boldsymbol{p}^k_{n_p}+\frac{1-\alpha }{||\mathbf{C}^k_{n_p}||_1}\cdot(\mathbf{C}^k_{n_p}\mathbf{Z}^k) ,
\end{equation}
here $\alpha\in[0,1]$ is the momentum coefficient. 
We leverage the fused CAM $\hat{\mathcal{M}}$ as the pseudo labels to update the class-wise positive prototype set $\mathcal{P}_\text{pos}$ and $1-\hat{\mathcal{M}}$ for the negative set $\mathcal{P}_\text{neg}$. Notably, each prototype is not involved in gradient backpropagation to avoid the noise from the classifier.

\noindent\textbf{Clue-based contrastive learning.} 
Based on the spatial perceptive clues, we introduce the clue-based contrastive learning~\cite{contra,contra2} to regularize the embedding space. More specifically, for each temporal dense embedding $\boldsymbol{z}^k_{n_k}$ and its most relative prototype $\boldsymbol{p}^k_{n_p}$, we adopt the cosine distance to measure their similarity: 
\begin{equation}
\mathtt{S}(\boldsymbol{z}^k_{n_k}, \boldsymbol{p}^k_{n_p})=\frac{\boldsymbol{z}^k_{n_k} \boldsymbol{p}^{k\top}_{n_p}}{\left\|\boldsymbol{z}^k_{n_k}\right\| \cdot\|\boldsymbol{p}^k_{n_p}\| / \tau},
\end{equation}
where $\tau$ indicates the temperature parameter. Subsequently, we enforce the pixel embedding to closely match its prototype and be distinct from other prototypes:
\begin{equation}
\mathcal{L}_\text{cbl}=\sum_k\sum_{n_k}\mathbbm{1}( \log (\sum_{\mathclap{{\boldsymbol{p} \in \mathcal{P}^{-}}}} \exp \mathtt{S}(\boldsymbol{z}^k_{n_k}, \boldsymbol{p})) - \mathtt{S}(\boldsymbol{z}^k_{n_k}, \boldsymbol{p}^k_{n_p}) ),
\end{equation}
where $\mathcal{P}^-=\{\mathcal{P}_\text{pos}\cup\mathcal{P}_\text{neg} \setminus \boldsymbol{p}^k_{n_p}\}$ and $\mathbbm{1}(\cdot)$ is an indicator function, being 1 if class $k$ appears in image-level labels and 0 otherwise. Minimizing the above objective can pull pixel embedding closely to the semantic center and push it away from other  illusory patterns, thereby sharpening the model decision boundary and mitigating the noise perturbation.

\subsection{Temporal-Aware Affinity Propagation}
\label{sec:affinity}
In this section, we propose the temporal-aware affinity propagation to cope with the wrong semantic bias arised from anomalous temporal clips.

\begin{figure}[t]
\centering
\includegraphics[width=1.0\linewidth]{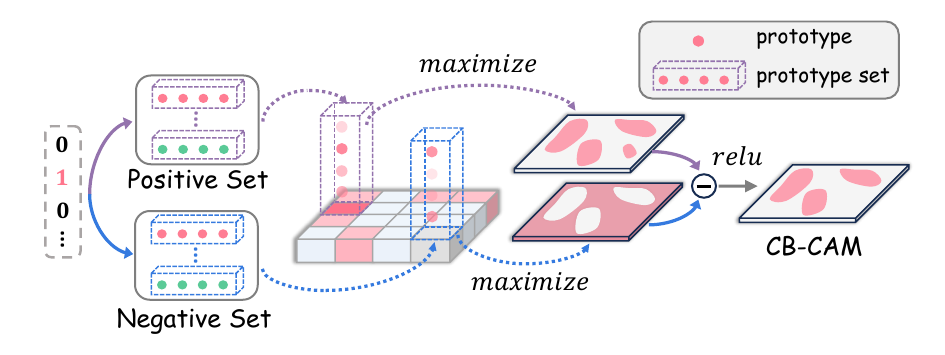}
\caption{\textbf{Visualization of the clue-based CAM generation.} This process is performed after training the classification network.}
\label{fig:lg}
\end{figure}

\noindent\textbf{Temporal-aware affinity mining.}
In the temporal encoder, the input tokens are normalized and projected into query matrix $\mathbf{Q}\in\mathbb{R}^{(K+T)\times d}$ and key matrix $\mathbf{K}\in\mathbb{R}^{(K+T)\times d}$ respectively. Then the self-attention $\mathcal{A}\in\mathbb{R}^{(K+T)\times(K+T)}$ with respect to $N_h\cdot N_w$ sequences is computed as below:
\begin{equation}
    \mathcal{A}=\mathtt{softmax}(\frac{\mathbf{Q}\mathbf{K}^\top}{\sqrt{d}}).
\end{equation}
Without additional computations nor supervision, we can explicitly extract the temporal-to-class attention $\tilde{\mathcal{A}}\in\mathbb{R}^{T\times K}$ from the self-attention $\mathcal{A}$, which represents the contributions of high-level representations to crop recognition at different temporal clips. We then reweight the the temporal sequence embeddings $\mathbf{Z}^\text{seq}_T$ based on the normalized attention $\tilde{\mathcal{A}}$:
\begin{equation}
   \mathcal{V}^k =\sum_{t=1}^{T} \tilde{\mathcal{A}}^k_t \cdot (\mathbf{Z}^\text{seq}_T)_{:,t,:}. 
\end{equation}
Intuitively, the modulated high-level representation $\mathcal{V}^k\in\mathbb{R}^{N_h\cdot N_w\times d}$ enlarges the variation among different semantic crops, thereby modeling more precise pixel relations.

\noindent\textbf{Affinity propagation and guidance.}
After obtaining the temporal-aware representation $\mathcal{V}$, we conduct affinity propagation to suppress erroneous semantic regions on the raw CAM. Particularly, the temporal-aware pairwise affinity of the pixel at position $i$ can be estimated as follows:
\begin{equation}
\mathop{\mathtt{Aff}(\boldsymbol{v}^k_i,\boldsymbol{v}^k_j)}\limits_{j \in \mathcal{N}_i} =\exp(\frac{\mathcal{S}(\boldsymbol{v}^k_i,\boldsymbol{v}^k_j)}{\sigma_i^k}),
\end{equation}
here $\sigma_i^k$ is the standard deviation of $\boldsymbol{v}_i^k$ and $\mathcal{N}_i$ indicates the local receptive fields (\eg, 8-way local neighbors~\cite{apro}). We denoise the raw CAM via iteratively propagating the pairwise affinity from the temporal-aware representation $\mathcal{V}$: 
\begin{equation}
	{\small \begin{split}
			\tilde{\mathcal M}^k_i =&\delta^k_i \sum_{j\in \mathcal{N}_i} \mathtt{Aff}(\boldsymbol{v}^k_i,\boldsymbol{v}^k_j) \cdot \mathcal{M}^k_j, \quad {\rm where}\\
			\quad \delta_i^k=&\frac{1}{\sum_{j\in\mathcal{N}_i}\mathtt{Aff}(\boldsymbol{v}^k_i,\boldsymbol{v}^k_j)}.
	\end{split}}
\end{equation} 
After obtaining the improved activation map, we align the raw CAM with it to effectively guide the learning process: 
\begin{equation}
 \mathcal{L}_\text{tap}= \sum_k \mathbbm{1} (|\tilde{\mathcal{M}^k} - \mathcal{M}^k |).
\end{equation}
By minimizing $\mathcal{L}_\text{tap}$, we can rectify erroneous activations in the raw CAM and incorporate temporal-aware priors into the embedding space, ultimately benefiting the perceptive clues.

\subsection{Network Optimization and Clue-based CAM Generation Strategy}
\label{sec:overall}
As shown in \Cref{fig:overview}, the whole objective function for training the classification network consists of four components: 
\begin{equation}
    \mathcal{L} = \mathcal{L}_\text{cls} + \mathcal{L}_\text{cls}^\text{aux} + \lambda_1\mathcal{L}_\text{cbl} + \lambda_2\mathcal{L}_\text{tap},
\end{equation}
here $\mathcal{L}_\text{cls}$ and $\mathcal{L}_\text{cls}^\text{aux}$ are the conventional binary cross entropy loss and $\lambda_i$ denotes the weight to rescale the loss terms. 

\noindent After training, we perform per-pixel perception based on the well-updated space-time clues in temporal dense embedding space to generate clue-based CAMs (CB-CAMs). For each embedding $\boldsymbol{z}_i$ in temporal dense embeddings $\mathbf{Z}_T^\text{dense}$, we measure the maximum similarity with the positive prototypes and minus the misguide activations with the negative prototypes:
\begin{equation}
 \mathcal{Y}_i^k = \mathtt{ReLu} (\max_{\boldsymbol{p}^+ \in\boldsymbol{p}_k^\text{pos}} \mathtt{S}(\boldsymbol{z}_i,\boldsymbol{p}^+) - \max_{\boldsymbol{p}^- \in\boldsymbol{p}_k^\text{neg}} \mathtt{S}(\boldsymbol{z}_i,\boldsymbol{p}^-)).
\end{equation}
If the class $k$ does not appear in a training image, we set $\mathcal{Y}^k$ to all zeros.
\Cref{fig:lg} gives a visualization of this generation process.
Finally, the CB-CAMs $\mathcal{Y}$ are filtered using a global background score to generate the pseudo labels, which are then used to train the SITS semantic segmentation network.

%% file: 04_exp.tex
\newcommand{\tablestyle}[2]{\setlength{\tabcolsep}{#1}\renewcommand{\arraystretch}{#2}\centering\footnotesize}
\newcommand{\reshl}[2]{
\textbf{#1} \fontsize{7.5pt}{1em}\textcolor[RGB]{0,150,0}{$\uparrow$ \textbf{#2}}
}

\begin{table*}[t]
  \centering
  \small
  \vspace{-1em}
  \subfloat[\parbox{12cm} {Pseudo labels performance of different methods on Germany and PASTIS \textit{train} set.}]
  {
    \hspace{-0.7em}
    \begin{minipage}[b]{0.58\linewidth}{

    \centering
    \tabcolsep=0.08cm
        \begin{tabular}{l|c|cc|cc}
        \toprule[1.2pt]
        \textbf{Method} & \textit{Type}  & \multicolumn{2}{c|}{\textit{Germany}~\cite{germany}} & \multicolumn{2}{c}{\textit{PASTIS}~\cite{pastis}} \\  
        \cmidrule{3-6}
         & & \textbf{OA} & \textbf{mIoU} & \textbf{OA} & \textbf{mIoU}
        \\
        \midrule
        MCTFormer{\tiny CVPR'22}~\cite{mctformer}& \multirow{6}{*}{$\mathcal{RGB}$} & 70.6 & 56.3 & 66.7& 49.6\\
        ViT-PCM{\tiny ECCV'22}~\cite{vitpcm}& & 66.8 & 52.2 & 69.3& 53.2\\ 
        LP-CAM{\tiny CVPR'23}~\cite{lpcam}&  & 64.7 & 49.1 & 67.0& 50.1\\
        TSCD{\tiny AAAI'23}~\cite{tscd}& & 68.4 & 52.7 & 67.2 & 51.3 \\ 
        DuPL{\tiny CVPR'24}~\cite{dupl}& & 66.2 & 50.6 & 65.5& 48.7\\
        SeCo{\tiny CVPR'24}~\cite{seco}& & 64.5 & 48.3 & 63.6 & 46.1 \\  
        \midrule 
         baseline &   & 82.5 & 74.1 & 81.2 & 69.5\\
         +PAMR{\tiny CVPR'20}~\cite{pamr}& \multirow{4}{*}{$\mathcal{SITS}$}  & 83.9 & 76.3 & 82.0 & 71.2 \\
         +TS-CAM{\tiny ICCV'21}~\cite{tscam}&    & 81.1 & 70.5 & 80.4 & 67.6 \\
         +SIPE{\tiny CVPR'22}~\cite{sipe} &   & 82.0 & 73.2 & 81.5 & 70.1\\
        +FPR{\tiny ICCV'23} ~\cite{fpr}&   & 82.4 & 75.6 & 81.7 & 71.0 \\
        \rowcolor{gray!15} 
        +ours-\textit{Exact}&    & \reshl{88.3}{5.8}& \reshl{80.6}{6.5}  & \reshl{84.1}{2.9}& \reshl{75.6}{6.1} 
        \\
        \bottomrule[1pt]
        \end{tabular}
    }
    \end{minipage}
    \label{tab:cam1}
  }
 \subfloat[Segmentation performance trained with pseudo labels gene- \\  rated by different methods on PASTIS \textit{test} set.]
  {
    \hspace{-5.5em}
    \begin{minipage}[b]{0.58\linewidth}{
    \begin{center}
    \tabcolsep=0.06cm
    \renewcommand\arraystretch{1.03}
    \begin{tabular}{l|c|ccc}
    \toprule[1.2pt]
    \textbf{Method} & \textit{Sup.} & \textbf{OA} & \textbf{mIoU} & \textit{ratio} \\  
    \midrule
    ConvLSTM~\cite{convlstm}  & \multirow{8}{*}{$\mathcal{M}$}   & 78.2& 50.1 \\
    BiCGRU~\cite{germany} &  & 80.5& 56.2 \\
    FPN-ConvLSTM~\cite{fpn_clstm}  &   & 81.9& 59.5 \\
    Unet-3D~\cite{unet2d} &   & 82.3 & 60.4 & -\\
    Unet-3Df~\cite{unet3df}&   & 82.1  & 60.2 \\
    U-TAE~\cite{pastis}&  & 83.2& 63.1\\
    TSViT~\cite{tsvit}&   & 83.4& 65.4 &100\% \\ 
    \midrule
    baseline & \multirow{6}{*}{$\mathcal{I}$} & 77.2& 57.8 &88\%\\
    +PAMR{\tiny CVPR'20}~\cite{pamr}& & 78.5 & 58.7 & 90\% \\ 
    +TS-CAM{\tiny ICCV'21}~\cite{tscam}&  &76.8& 56.5 & 86\% \\
    +SIPE{\tiny CVPR'22}~\cite{sipe}& & 77.1 & 58.0 & 89\% \\
    +FPR{\tiny ICCV'23} ~\cite{fpr}& & 78.2 & 58.4 & 89\% \\
    \rowcolor{gray!15} +ours-\textit{Exact}&   & \reshl{80.2}{3.0} & \reshl{62.0}{4.2} & \textcolor[RGB]{0,150,0}{\textbf{95\%}} \\
    \bottomrule[1pt]
    \end{tabular}
    \end{center}
    }
    \end{minipage}
    \label{tab:seg}
  }
    \caption{\textbf{Comparisons with existing WSSS methods in OA(\%) and mIoU(\%).} 
    $\mathcal{RGB}$: the networks designed for processing natural images. $\mathcal{SITS}$: the networks designed for processing SITS (include four adapted modules from WSSS methods in natural image domain). $\mathcal{M}$: the SITS segmentation networks supervised by pixel-level masks. $\mathcal{I}$: the TSViT segmentation network supervised by pseudo labels from different methods. The \textit{ratio} refers to the proportion of mIoU between fully supervised and weakly supervised of TSViT segmentation. }
  \vspace{-0.5em}
\end{table*}%

\section{Experiments}
\label{sec:exp}
\vspace{-0.3em}
\subsection{Experimental Setup}
\vspace{-0.5em}
\noindent\textbf{Datasets \& evaluation metric.}~
We conducted comprehensive experiments on two widely used Satellite-2 time series crop recognition datasets to validate the effectiveness of our approach quantitatively. 
The \textit{PASTIS} dataset~\cite{pastis} contains $2433$ multi-spectral time series of size $128\times 128$, each series including $10$ bands and $33$ to $61$ temporal observations. It consists of $18$ crop types and a background category. Following the settings of~\cite{tsvit}, we used the fold-1 among the five folds provided in PASTIS. We partitioned each sample into multiple patches and assigned category labels according to the mask annotations (see the supplement for details).
The \textit{Germany} dataset~\cite{germany} comprises $137k$ field parcels of time series imagery with $13$ spectral bands. Each series contains $36$ observations and is labeled with $17$ crop types. We employed the mean Intersection over Union (mIoU) and pixel-wise overall accuracy (OA) as evaluation metrics, both widely used to measure segmentation performance.

\noindent\textbf{Implementation details.}~The classification network was trained on 8 NVIDIA RTX 3090 GPUs with batch size 8 for 15k iterations. During the training stage, we used the AdamW optimizer~\cite{adamw} with an initial learning rate of 1e-3 and cosine weight decay~\cite{sgdr}. 
As the raw CAM were unreliable in early iterations, the gradients of $\mathcal{L}_\text{cbl}$ were backpropagated after the 4k iteration. For the raw CAM filtering, the threshold score $(\mu_l,\mu_h)$ was set to $(0.2,0.4)$. Due to the compact intra-class patterns, we set the number of class-specific prototypes $N_p$ to $2$, and the momentum used to update the prototypes was set to $0.999$. Moreover, the other hyper-parameters $\eta$, $\lambda_1$, $\lambda_2$ and $\tau$ were empirically set to $0.05$, $0.01$, $0.015$ and $0.1$, respectively.
We employed the original TSViT with a segmentation decoder as the semantic segmentation model for SITS. The training details of the segmentation model exactly followed the settings in~\cite{tsvit} without any modifications. Please refer to supplementary materials for more details.

\subsection{Experimental Results}
\label{sec:4.2}

\vspace{-0.2em}
\noindent\textbf{Base Architectures.} 
In the evaluation stage, we fuse the raw CAMs generated by the spatial dense embeddings $\mathbf{Z}_S^\text{dense}$ and temporal dense embeddings $\mathbf{Z}_T^\text{dense}$ as our baseline.  

\vspace{-0.25em}

\subsubsection{Comparison with other WSSS Methods on CAMs.}
The key of the weakly supervised learning is to provide reliable training pseudo labels for segmentation network. Thus, we conducted comparative experiments on the \textit{train} splits to validate the quality of pseudo labels.

\noindent\textbf{Competing methods adaptation.} We reimplemented six well-performed transformer-based WSSS methods designed for natural images as the competing methods. To accommodate the temporal dimension, we transformed the input dimensions of SITS into a 3D format $(T\times C, H,W)$ for these methods.
In addition, to enable a fairer comparison between \textit{Exact} and existing WSSS works, we carefully selected four off-the-shelf modules from other WSSS methods in the natural image domain and adapted them to the temporal-spatio framework. These modules were also reimplemented within the temporal dense embedding space to align with our approach. We attempted various adaptation strategies, more details and analysis available in supplementary materials.

\begin{table}[t]
\centering
\tabcolsep=0.18cm 
\small 
\renewcommand\arraystretch{1.2} 
\begin{tabular}{l|ccc|ccc}
\toprule[1pt]
\textbf{Method} & $\mathcal{L}_\text{cbl}$ & $\mathcal{L}_\text{tap}$ &    \textit{CB-CAM} & Recall  & \textbf{OA} & \textbf{mIoU} \\  
\midrule
baseline   &  &  &  & 81.1 & 81.2 & 69.5 \\
 &   & \CheckmarkBold & & 81.8 & 82.4  & 72.3  \\
 & \CheckmarkBold  & \CheckmarkBold & & 82.5 & 83.0 & 73.4 \\
\rowcolor{gray!15} ours-\textit{Exact} & \CheckmarkBold   & \CheckmarkBold &\CheckmarkBold & \textbf{84.8} & \textbf{84.1} & \textbf{75.6} \\
\bottomrule[1pt]
\end{tabular}
\caption{\textbf{Ablation study results of different components.} The baseline is a setting with only image-level classification loss.}
\label{tab:ablation}
\end{table}

\begin{table}[t]
\centering
\tabcolsep=0.13cm 
\small 
\renewcommand\arraystretch{1.2} 
\begin{tabular}{l|cc|cccc}
\toprule[1pt]
\textbf{Method} &\textit{Spa.} & \textit{Tem.} & Recall & Precision  & \textbf{OA} & \textbf{mIoU} \\  
\midrule
 & \CheckmarkBold & & \textbf{81.9}  & 78.4 &  80.5 & 66.7 \\
 &  & \CheckmarkBold & 80.8  & 81.6  &   81.0  & 68.3  \\
\rowcolor{gray!15} baseline CAM & \CheckmarkBold & \CheckmarkBold & 81.1 & \textbf{83.5} & \textbf{81.2} & \textbf{69.5} \\
\bottomrule[1pt]
\end{tabular}
\caption{\textbf{Baseline CAMs from different embedding spaces.}  Overall, the fused CAM shows the best performance.}
\label{tab:ablation2}
\end{table}

\begin{table}[t]
\centering
\tabcolsep=0.18cm 
\small 
\renewcommand\arraystretch{1.2} 
\begin{tabular}{l|cc|cccc}
\toprule[1pt]
\textbf{Method} &\textit{Low.} & \textit{Ta.} & Recall & Precision  & \textbf{OA} & \textbf{mIoU} \\  
\midrule
baseline   &  &  & 81.1 & 83.5 & 81.2 & 69.5 \\
  & \CheckmarkBold & & 80.9 & 84.0  &  81.1 & 69.9 \\
\rowcolor{gray!15} ours-\textit{Exact} &  & \CheckmarkBold & \textbf{81.8} & \textbf{85.6} &  \textbf{82.4}  & \textbf{72.3}  \\
   & \CheckmarkBold & \CheckmarkBold & 81.7 & 85.2 & 82.0 & 71.8 \\
\bottomrule[1pt]
\end{tabular}
\caption{\textbf{Impact of modeling affinity from different sources.} The temporal-aware affinity remarkably reduces noise in low-level cues.}
\label{tab:ablation3}
\end{table}

\noindent\textbf{Quantitative results.} 
\cref{tab:cam1} presents the mIoU and OA of the pseudo labels generated by other WSSS methods and our \textit{Exact}. As can be seen, the WSSS methods designed for natural images only achieve a maximum 53.2\% mIoU (refer to PASTIS) with the limited encoding capability for SITS. Furthermore, due to the distinct data characteristics, the off-the-shelf modules struggle to perform well on SITS, and may even disturb the learning process. These phenomena indicate that the advancements in natural image cannot be directly translated to benefit the SITS domain. By contrast, our \textit{Exact} overcome the intrinsic challenges from SITS, delivering the best performance compared to others.

\subsubsection{Segmentation results trained by pseudo labels.}
After obtaining the pseudo labels, we use them as replacement for ground truth to train the SITS semantic segmentation network. We employ TSViT followed by a segmentation head as our segmentation model, which is the top-performing approach in fully supervised oracles. 

\noindent\textbf{Quantitative results.} In \cref{tab:seg}, we show that the semantic segmentation model can achieve 95\% to its fully supervised performance (refer to mIoU) using pseudo labels generated by \textit{Exact}. Our best results are 80.2\% OA and 62.0\% mIoU on PASTIS \textit{test} set, which surpasses the baseline by 3.0\% and 4.2\%, respectively.
The experiment results quantitatively imply that the labels quality outperforms other methods by a large margin. In contrast, WSSS networks in natural image have undergone a series of evolutions to barely achieve 90\% of fully supervised performance. This suggests that SITS WSSS techniques offers greater potential in SITS scenario.

\begin{figure}[t]
\centering
\includegraphics[width=1.0\linewidth]{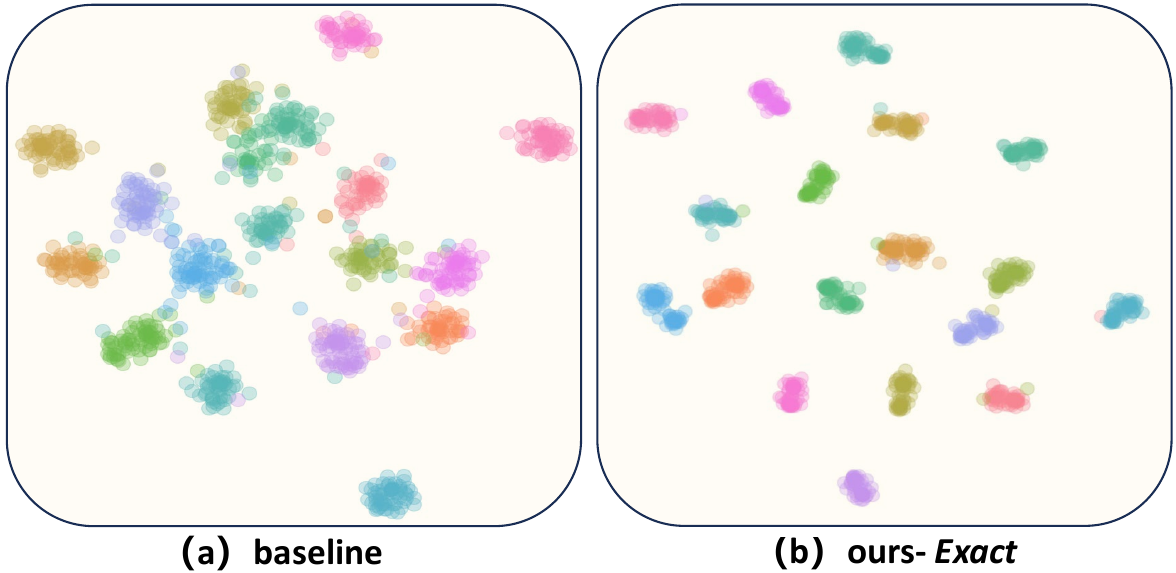}
\caption{\textbf{Visualization of temporal feature spaces on PASTIS \textit{train} set.} The feature dimensions are reduced by t-SNE~\cite{tsne}.}
\label{fig:feat}
\end{figure}

\begin{figure}[t]
\centering
\includegraphics[width=1.0\linewidth]{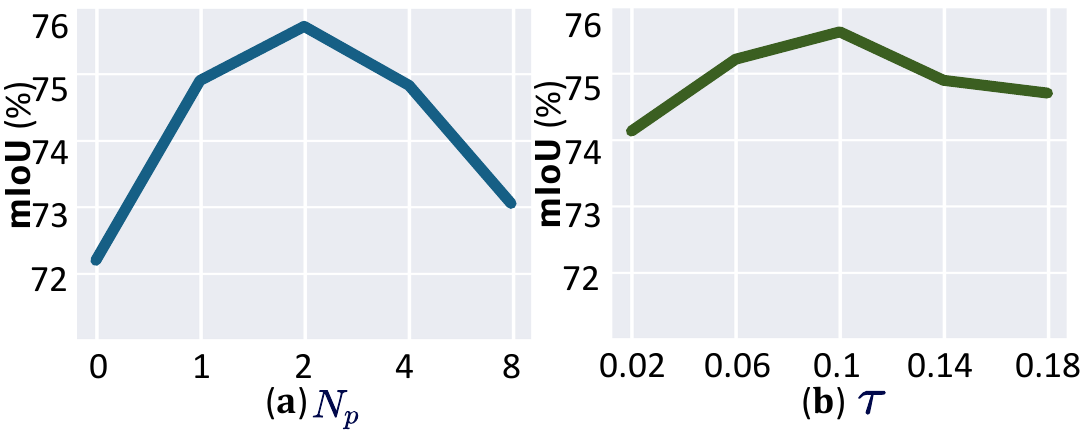}
\caption{\textbf{Effect of the hyper-parameters.} (a) the number of class-specific prototypes $N_p$. (b) the temperature of similarity $\tau$.}
\label{fig:hyper}
\end{figure}

\vspace{-0.1cm}
\subsection{Ablation Studies}
\label{sec:4.3}
\vspace{-0.08cm}
In this section, our primary purpose is to demonstrate the collective effectiveness of all components within our approach. We also choose the fused raw CAM as the baseline, and the CAM comparison results reported on PASTIS \textit{train} set.

\noindent\textbf{All the components matter.} Our \textit{Exact} consists of several components, including spatial perceptive clues exploration, temporal-aware affinity propagation and clue-based CAM generation strategy. We validate the contribution of each module, the results are presented in \cref{tab:ablation}. 
We can see that the clue-based contrastive learning $\mathcal{L}_\text{cbl}$ and the temporal-aware affinity propagation $\mathcal{L}_\text{tap}$ improves the performance by 1.8\% OA and 3.9\% mIoU, respectively. This suggests that these two modules complement each other, regularizing the entire embedding space and ultimately sharpening the decision boundary. We provide an intuitive comparison in \cref{fig:feat}, after introducing the both objectives, the intra-class features become more compact while the inter-class features are more separated. Besides, using the well-updated space-time clues to generate CAM (\ie, CB-CAM) within the temporal dense embedding space brings a further improvement of 2.3\% recall, 1.1\% OA and 2.2\% mIoU.

\begin{figure*}[t]
\centering
\includegraphics[width=1.0\linewidth]{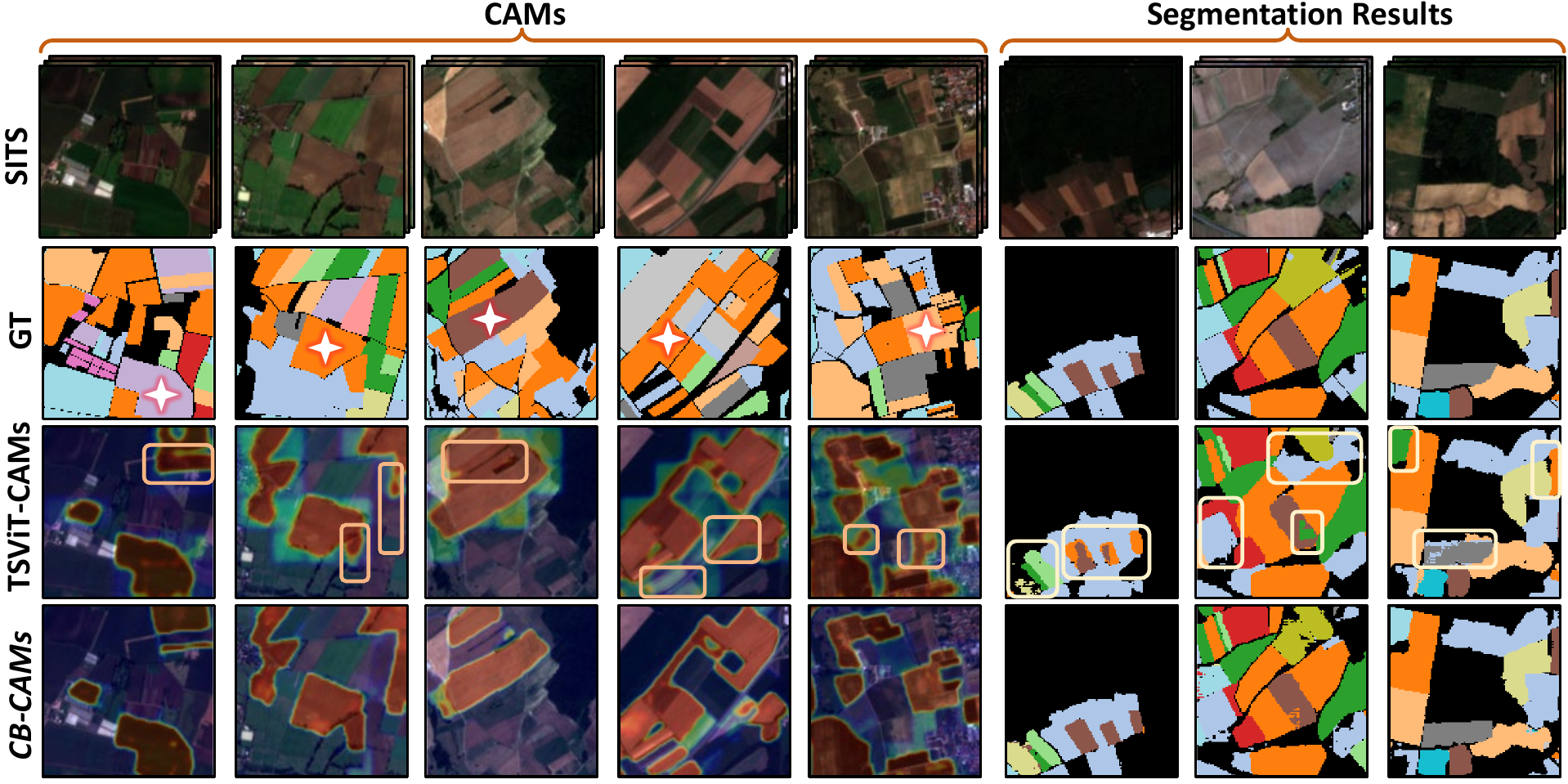}
\caption{\textbf{Qualitative results between baseline TSViT-CAM and CB-CAMs derived by \textit{Exact} on PASTIS dataset.} \textbf{Left}: CAMs comparisons. \textbf{Right}: Semantic segmentation comparison results. 
The stars represent the corresponding activation category.
}
\label{fig:cams}
\end{figure*}

\noindent\textbf{Baseline CAM from different embedding spaces.}
In \cref{tab:ablation2}, we investigate the performance of the raw CAMs derived from different dense embedding spaces. The raw CAM obtained from the spatial dense embeddings $\mathbf{Z}_S^\text{dense}$ falls significantly behind that obtained from the temporal dense embeddings $\mathbf{Z}_T^\text{dense}$ in terms of precision, OA and mIoU. It demonstrates that the temporal correlations contain more critical information compared to the spatial context in this task. We combine the CAMs from both embedding spaces as our baseline (the last row in \cref{tab:ablation2}) and utilize the resulting pseudo labels to guide spatial clustering. 

\noindent\textbf{Effect of the temporal-aware affinity.} Mining low-level cues affinity (\eg, color and intensity) as additional guidance is prevalent in natural images. We also attempted to incorporate the low-level cues to model pairwise affinity in SITS, the results can be found in \cref{tab:ablation3}. Compared to our proposed temporal-aware affinity, modeling low-level cues offers only limited improvement. This is due to the presence of substantial noise interference within the low-level cues of SITS (\eg, cloud cover and shadow). In contrast, our approach highlights the contributions of pivotal temporal clips within a high-level embedding space, effectively mitigating the wrong semantic bias.

\noindent\textbf{Effect of the hyper-parameters.} 
\cref{fig:hyper} shows the effects of the prototype number $N_p$ and the temperature $\tau$. Notably, utilizing only a minimal number of prototypes (\ie, $N_p=2$) yields optimal result, whereas increasing the number of prototypes beyond this point brings negative impact on the model. This indicates that the excessive number may enforce prototypes to focus on local discriminative patterns within the class, leading to under-activation issue (the major difficulty in natural image).
More analysis of parameters is specifically discussed in supplementary materials.
\vspace{-0.1cm}
\subsection{Qualitative Results}
\vspace{-0.1cm}
\cref{fig:cams} presents the visual comparison between the baseline TSViT-CAMs and the proposed CB-CAMs generated by  \textit{Exact} on PASTIS dataset. The first five columns show the visualization of the CAMs. As we can see, our method remarkably suppress the erroneous regions, whose shape is closer to ground truth masks than baseline. Our high-quality pseudo labels subsequently enhance the segmentation performance compared to the baseline, as shown in last three columns. However, \textit{Exact} still exhibits limitations in edge processing, which will be the focus in our future work.

%% file: 10_conclusion.tex
\vspace{-0.1cm}
\section{Conclusion and Broader Impact}
\label{sec:conclusion}

\vspace{-0.1cm}
In this paper, we propose a tailored WSSS approach \textit{Exact} to alleviate the daunting annotation challenge in SITS crop mapping task. 
\textit{Exact} explores space-time perceptive clues to capture the essential patterns of different crop types, thereby overcoming the issues of spatial noise perturbation and wrong temporal semantic bias.
Extensive experiments show the superiority of \textit{Exact}-generated masks both quantitatively and qualitatively.
We believe that our method marks a pioneering step in effectively applying WSSS technologies to SITS. 
If follow-up work can find and resolve the limitations under our framework, there is great potential that only image-level labels are needed for crop mapping in the future.

%% file: 12_appendix.tex
\pagestyle{empty}

In this supplementary material, we provide the following contents: 1) more implementation details of dataset, our \textit{Exact} and competing methods ($\S$\ref{sec:a1}), 2) the difference with the previous prototype-based methods ($\S$\ref{sec:a2}), 3) additional experimental results and analyses ($\S$\ref{sec:a3}).

\section{Additional Implementation Details}
\label{sec:a1}

\subsection{Dataset and Model Design}
\textbf{Data Setting.} Since the multi-class labels are absent in original PASTIS~\cite{pastis} and Germany~\cite{germany} datasets, we employ the following strategy to assign multi-class labels to each SITS: If the number of masks for class $k$ constitutes at least 1\% of the spatial size, the class $k$ is deemed to be present in SITS. To accommodate a large set of experiments, we only use fold-1 among the five folds provided in PASTIS. 

\noindent\textbf{Model Setting.} Following the settings in~\cite{tsvit}, we set the vector dimension $d$ to 128. The temporal encoder and spatial encoder comprise 8 and 4 layers, respectively. The spatial size of each patch is set to $h=w=2$. 
For the spatial clustering, we apply $\ell_2$ normalization to the pixel embedding before measuring the similarity with prototypes. 
For the temporal-aware affinity mining, we normalize the temporal-to-class attention $\tilde{\mathcal{A}}$ to the range of [0,1] using $\mathtt{softmax}$ function and subsequently reweight the temporal sequence embeddings by the normalized attention. 
Besides, we iteratively propagate the temporal-aware pairwise affinity among neighboring pixels, with the iterations are set to 3. To derive the final pseudo labels, we apply a global threshold (\ie, $\theta_\text{bg}=0.3$) to separate the foreground and background in the CB-CAMs $\mathcal{Y}$, as following~\cite{mctformer,seam}.

\subsection{Competing Methods and Modules}
\subsubsection{Adaptation of Competing Methods} 
For WSSS methods originally designed for natural images, we attempt several adaptation strategies to enable their application to SITS inputs, as follows:

\noindent\textbf{Strategy 1.} 
Given a SITS with dimensions $[T,C,H,W]$, we partition it into $T$ single-temporal samples, each with dimensions $[C+1, H, W]$. The additional channel represents the temporal position embedding, which is used to differentiate distinct temporal samples. 
We then check each temporal sample for cloud cover by evaluating the maximum signal intensity, and removing any samples identified as \textit{cloudy}. Each remaining temporal sample is individually processed by the WSSS networks to generate CAM. Finally, we average the CAMs from temporal samples to obtain a single CAM for SITS. The pseudo-code is attached in \cref{alg:strategy1}.

\noindent\textbf{Strategy 2.} 
We reconstruct the SITS into the 3D format $[T\times C,H,W]$ by merging the first and second dimensions. Subsequently, we directly input the reconstructed data into the WSSS model to obtain the CAM for SITS.

\begin{algorithm}[tb]
    \renewcommand{\thealgorithm}{A}
    \caption{Adaptation strategy 1.}
    \label{alg:strategy1}
    \textbf{Input}: SITS $\boldsymbol I\in \mathbb{R}^{T \times C \times H \times W}$.\\
    \textbf{Parameter}: cloud threshold $thr$. \\
    \textbf{Function}: WSSS network $f$ designed for natural images. \\
    \textbf{Output}:single CAM $\mathcal{M} \in \mathbb{R}^{N_h\cdot N_w\times K} $ for SITS.\\
    \begin{algorithmic}[1] 
        \STATE Initialize CAMs list $\mathcal{M}^\prime \leftarrow  [\ ]$;
        \FOR{$t \leftarrow 1$ \TO $T$}
        \IF {$\mathtt{max}(\boldsymbol I_{t})<thr$} 
        \STATE {\hfill \footnotesize \texttt {$\rhd$ cloud cover check}}\\
         $\boldsymbol I_{t} \leftarrow \mathtt{concat}(\boldsymbol I_t, \ \text{time\  position})$ 
        \STATE {\hfill \footnotesize \texttt {$\rhd$ single temporal sequence input}}
        \STATE $ \mathcal{M}_{t} \leftarrow f (\boldsymbol I_{t})$;\\
        \STATE $ \mathcal{M}^\prime \leftarrow \mathcal{M}^\prime.\mathtt{append}(\mathcal{M}_{t})$;\\
        \ENDIF
        \ENDFOR
        \STATE $\mathcal{M} \leftarrow \mathtt{mean}(\mathcal{M}^\prime)$ {\hfill \footnotesize \texttt {$\rhd$ Average over $\texttt{T}$ dimension}}
        \RETURN $ \mathcal{M} $;
    \end{algorithmic}
\end{algorithm}

\subsubsection{Adaptation of Competing Modules}
Most of modules designed in natural image WSSS cannot be applied in SITS scenario due to the distinct data processing pipeline. We carefully select and reimplement four modules that can be adapted to temporal-spatio network.

\noindent\textbf{PAMR}~\cite{pamr}. For this method, we incorporate the nGWP and PAMR modules. We feed the score map output from temporal encoder and spatial encoder into the nGWP module to replace the convenient GAP layer. The PAMR module is widely utilized in natural image WSSS. Here, we follow the common practice that using low-level intensities as the input to PAMR to refine the fused raw CAM. 

\noindent\textbf{TS-CAM}~\cite{tscam}. Since this method performs semantic re-allocation and semantic aggregation from a spatial perspective, we follow the original settings in ~\cite{tscam} that compute TS-CAM within the spatial encoder. Besides, we replace the multi-class tokens in spatial encoder with single-class token to maintain consistency with the original implementation.

\noindent\textbf{SIPE}~\cite{sipe}. This model computes the inter-pixel semantic correlations in feature space, providing additional guidance for extending the CAM. We reproduce SIPE in the temporal embedding space to align with our proposed module.

\noindent\textbf{FPR}~\cite{fpr}. We calculate the region-level contrast loss and pixel-level rectification loss proposed by FPR in the temporal embedding space. Note that both SIPE and FPR use prototypes, and we set the number of prototypes per class to 2 to better match the inherent characteristics of the SITS.

\begin{table}[t]
\small
\centering
\renewcommand{\arraystretch}{1.05}
\setlength{\tabcolsep}{2.5mm}
\begin{tabular}{l|cc|>{\columncolor{gray!15}}c>{\columncolor{gray!15}}c}
\toprule[1.2pt]
\textbf{Method} & \multicolumn{2}{c|}{\textit{Strategy} 1} & \multicolumn{2}{c}{\cellcolor{gray!15}\textit{Strategy} 2}  \\ 
    \cmidrule{2-5}
    & \textbf{OA}  &  \textbf{mIoU} & \textbf{OA}  & \textbf{mIoU}      \\
    \midrule
MCTFormer{\tiny CVPR'22}~\cite{mctformer} & 63.8 & 41.5 & 66.7 & 49.6        \\ 
ViT-PCM{\tiny ECCV'22}~\cite{vitpcm}& 65.1 & 46.3 & 69.3& 53.2 \\
TSCD{\tiny AAAI'23}~\cite{tscd}& 64.8 & 45.9 & 67.2 & 51.3 \\ 
DuPL{\tiny CVPR'24}~\cite{dupl}& 63.4 & 40.7 & 65.5& 48.7\\

\bottomrule[1.2pt]
\end{tabular}
\vspace{-3pt}
\caption{\textbf{The performance of WSSS designed for natural images under different adaptation strategies.}}
\label{tab:wsss}
\end{table}

\begin{figure}[t]
\centering
\includegraphics[width=1.0\linewidth]{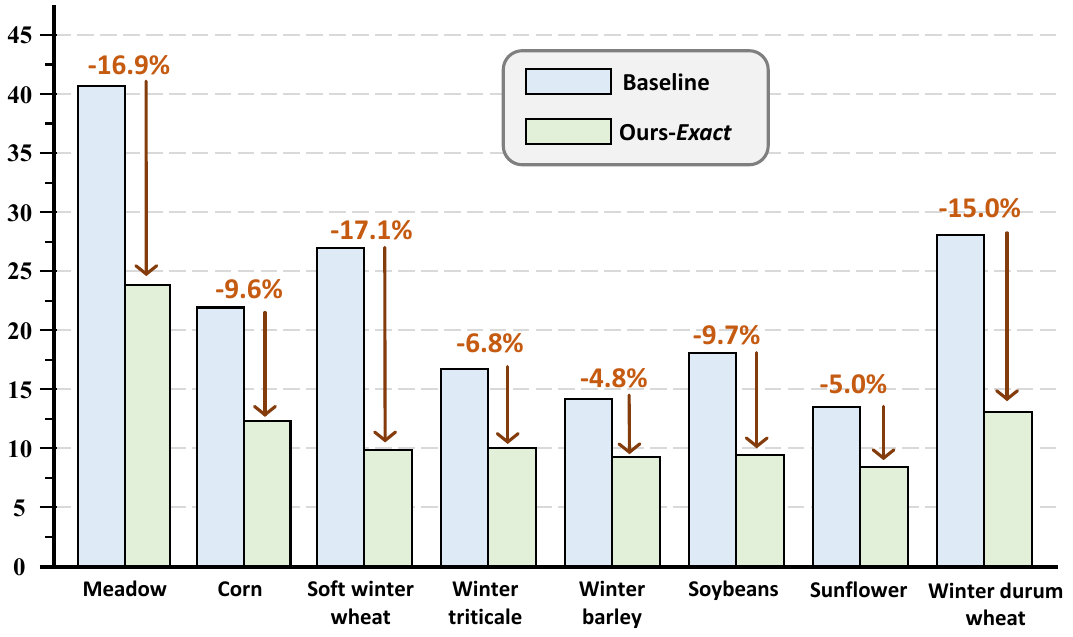}
\caption{\textbf{False discovery rate (FDR) of baseline and \textit{Exact}.} The results are evaluated on the PASTIS \textit{train} set for several major crop types. \textit{Exact} significantly reduce the FDR across different crops. }
\label{fig:fpr}
\end{figure}

\section{Remarks on difference with previous works in natural images.}
\label{sec:a2}
\textbf{Existing prototype-based WSSS works.} The prototype-based methods have been explored in the WSSS for natural images. Existing prototype-based WSSS works on natural images primarily focus on the following aspects: 1) setting a large number of prototypes (approximately 30 per class) to capture diverse intra-class patterns and leveraging these prototypes to reduce intra-class variation~\cite{seco,rca,fpr}. 2) performing clustering on the batch-level and utilize the prototypes obtained from the current batch to expand the raw CAM~\cite{sipe,ppc,lpcam}.

\noindent\textbf{Difference with the previous works in natural images.} Our approach is different from existing prototype-based WSSS methods in the following ways: 1) our objective is to explicitly capture compact intra-class patterns using a minimal number of prototypes ($N_p=2$) and leverage the semantics to enlarge the variations across different crops.
2) we perform momentum updating at the dataset-level and conduct spatial clustering in the most class-relative regions. 3) we discard the classifier weights and directly utilize the well-updated prototypes to generate the final CAMs. 

Our method thoroughly considers the unique characteristics of SITS and introduces tailored strategies, thereby achieving impressive performance. By contrast, prototype-based methods designed for natural images yield only limited improvements. As shown in the last column of Tab.\textcolor[RGB]{42,120,189}{1a} in the main paper, SIPE~\cite{sipe} and FPR~\cite{fpr} (two prototype-based methods for natural images) bring only 0.6\% and 1.5\% improvements compared to baseline, while our \textit{Exact} achieves a substantial 6.1\% enhancement.

\section{Additional Experiments and Analyses}
\label{sec:a3}
Due to constraints on page space, we are unable to present all experimental results within the main paper. In this section, we provide more experimental results and analyses both quantitatively and qualitatively to support the main paper.

\subsection{Different Adaptation Strategies.} 

We evaluate the performance of the WSSS under different adaptation strategies on PASTIS \textit{train} set, the results are shown in \cref{tab:wsss}. 
It can be observed that the WSSS models generally perform better under \textit{Strategy} 2. This is because merging the temporal dimensions during input allows the model to implicitly focus on pivotal temporal clips, thereby mitigating the adverse effects of anomalous temporal periods to some extent. 
To eliminate the influence of irrelevant factors, we choose the \textit{Strategy} 2 in the main paper to reimplement the WSSS method designed for natural images.

\begin{table}[t]
\small
\centering
\renewcommand{\arraystretch}{1.05}
\setlength{\tabcolsep}{0.7mm}
 \begin{tabular}{l|c|cccc}
    \toprule[1.2pt]
    \textbf{Method} & \textit{Sup.} & \textbf{OA} & \textbf{mIoU} & O.\textit{ratio} & m.\textit{ratio} \\  
    \midrule
    TSViT~\cite{tsvit}& $\mathcal{M}$ & 95.0 & 84.8 & 100\% & 100\% \\
    \midrule
    baseline & \multirow{6}{*}{$\mathcal{I}$} & 84.7 & 73.6 & 89\%  & 87\% \\
    +PAMR{\tiny CVPR'20}~\cite{pamr} & & 85.4 & 74.8 &  90\% & 88\% \\
    +TS-CAM{\tiny ICCV'21}~\cite{tscam}& & 82.6 & 71.9  & 87\% & 85\% \\
    +SIPE{\tiny CVPR'22}~\cite{sipe}& & 84.2 & 73.1   & 89\%  & 86\% \\
    +FPR{\tiny ICCV'23} ~\cite{fpr}& & 84.5 & 73.9  & 89\%  & 87\% \\
    \rowcolor{gray!15} +ours-\textit{Exact}&   & \reshl{90.1}{5.4} & \reshl{79.9}{6.3} & \textcolor[RGB]{0,150,0}{\textbf{95\%}} &\textcolor[RGB]{0,150,0}{\textbf{94\%}} \\
    \bottomrule[1pt]
    \end{tabular}
\vspace{-3pt}
\caption{\textbf{The TSViT segmentation network performance trained with pseudo labels on Germany \textit{test} set.} All pseudo labels are consistent with those described in the main paper. O.\textit{ratio} and m.\textit{ratio} refer to the proportion of OA and mIoU between weakly supervised and fully supervised of segmentation performance. }
\label{tab:germany}
\end{table}

\begin{table}[t]
\small
\centering
\renewcommand{\arraystretch}{1.05}
\setlength{\tabcolsep}{0.7mm}
 \begin{tabular}{l|c|cccc}
    \toprule[1.2pt]
    \textbf{Method} & \textit{Sup.} & \textbf{OA} & \textbf{mIoU} & O.\textit{ratio} & m.\textit{ratio} \\  
    \midrule
    U-TAE~\cite{pastis}& $\mathcal{M}$ & 83.2& 63.1 & 100\% & 100\% \\
    \midrule
    baseline & \multirow{6}{*}{$\mathcal{I}$} & 76.1 & 55.8 & 91\%  & 88\% \\
    +PAMR{\tiny CVPR'20}~\cite{pamr}&  & 77.5 & 56.7 & 93\% & 90\% \\ 
    +TS-CAM{\tiny ICCV'21}~\cite{tscam}&  & 75.3 & 54.2 & 90\% & 86\% \\
    +SIPE{\tiny CVPR'22}~\cite{sipe}& & 76.4 & 56.1 & 92\%  & 89\% \\
    +FPR{\tiny ICCV'23} ~\cite{fpr}& & 76.8 & 56.5 & 92\%  & 90\% \\
    \rowcolor{gray!15} +ours-\textit{Exact}&   & \reshl{82.2}{6.1} & \reshl{60.9}{5.1} & \textcolor[RGB]{0,150,0}{\textbf{99\%}} &\textcolor[RGB]{0,150,0}{\textbf{96\%}} \\
    \bottomrule[1pt]
    \end{tabular}
\vspace{-3pt}
\caption{\textbf{The U-TAE segmentation network performance trained with pseudo labels on PASTIS \textit{test} set.} All pseudo labels are consistent with those described in the main paper.}
\label{tab:utae}
\end{table}

\begin{table}[t]
\centering
\tabcolsep=0.18cm 
\small 
\renewcommand\arraystretch{1.2} 
\begin{tabular}{l|ccc|cc}
\toprule[1pt]
\textbf{Method} & $\mathcal{L}_\text{cbl}$ & $\mathcal{L}_\text{tap}$ &    \textit{CB-CAM} & \textbf{OA} & \textbf{mIoU} \\  
\midrule
baseline   &  &  &   & 81.2 & 69.5 \\
 &  \CheckmarkBold &  &  & 81.9  & 71.6  \\
 &   & \CheckmarkBold &  & 82.4  & 72.3  \\
 &  &  & \CheckmarkBold & 82.2  & 72.5  \\
  & \CheckmarkBold  & \CheckmarkBold &  & 83.0 & 73.4 \\
  & \CheckmarkBold  &  & \CheckmarkBold & 82.8  & 73.3  \\
 &  & \CheckmarkBold & \CheckmarkBold & 83.2 & 73.8 \\
\rowcolor{gray!15} ours-\textit{Exact} & \CheckmarkBold   & \CheckmarkBold &\CheckmarkBold  & \textbf{84.1} & \textbf{75.6} \\
\bottomrule[1pt]
\end{tabular}
\caption{\textbf{Additional ablation results of different components.}}
\label{tab:ablation}
\end{table}

\subsection{More Comparisons}
\noindent\textbf{Effect of our method on correcting false positives.}
We evaluate the false discovery rate (FDR) for each category of pseudo-labels to quantitatively analyze the effectiveness of our method in mitigating the over-activation regions. The FDR can be computed as follows: 
\begin{equation}
\texttt{FDR} = \frac{\texttt{FP}}{\texttt{FP} + \texttt{TP}}
\end{equation}
here \texttt{FP} and \texttt{TP} denote the number of false positive and true positive pixel pseudo labels for each class.
We compare the FDR of pseudo labels generated by TSViT-CAMs (baseline) and ours \textit{Exact}, the results are shown in \cref{fig:fpr}. It can be seen that due to the noise perturbations, the baseline exhibits more false positive pixels, leading to inferior perceive ability. Our method significantly suppresses erroneous activation regions and reduces the FDR across different crops, thereby delineating the crop regions more precisely.

\noindent\textbf{Segmentation results on Germany dataset.} We further validate the performance of segmentation network trained by different pseudo labels on the Germany~\cite{germany} dataset. As in the main paper, we employ the original TSViT~\cite{tsvit} with a segmentation decoder as our SITS semantic segmentation network. As shown in \cref{tab:germany}, training the segmentation network with \textit{Exact}-generated labels achieves the best results, improving the baseline by 5.4\% in OA and 6.3\% in mIoU, respectively. This indicates that our method can show consistently superior performance across various SITS crop mapping benchmarks.

\noindent\textbf{Segmentation results for other SITS segmentation network.}
To further demonstrate the superiority of our method, we replace the TSViT with the U-TAE~\cite{pastis} as our semantic segmentation network and evaluate its performance under various pseudo labels generated by different methods. The results are shown in \cref{tab:utae}. Notably, using the pseudo labels generated by \textit{Exact}, U-TAE can achieve 99\% and 96\% of the fully supervised OA and mIoU respectively, showcasing the impressive performance of our method. These findings indicate that training lightweight network with the pseudo labels generated by our method has the potential to achieve performance comparable to its fully supervised paradigm.

\begin{table}[t]
\small
\centering
\begin{subfigure}{0.235\textwidth}
\renewcommand{\arraystretch}{1.03}
\setlength{\tabcolsep}{1.7mm}
\centering
\begin{subtable}[t]{1.0\textwidth}
\begin{tabular}{cc|cc}
\toprule[1.2pt]
$\mu_l$          &  $\mu_h$     & \textbf{OA}                         & \textbf{mIoU}                           \\ \midrule
0.15   & 0.40  & 83.8     & 75.2   \\
0.20   &  0.35  &  83.2  &  75.0   \\
\rowcolor{gray!15}
{\textbf{0.20}} & {\textbf{0.40}} & {\textbf{84.1}} & {\textbf{75.6}} \\
0.20   & 0.45  & 83.3     & 75.1   \\
0.25   & 0.40  & 83.0    & 75.3
\\
\bottomrule[1.2pt]
\end{tabular}
\vspace{2pt}
\caption{Filtering thresholds $\mu$.}
\label{tab:score}
\end{subtable}
\end{subfigure}
\begin{subfigure}{0.235\textwidth}
\renewcommand{\arraystretch}{1.03}
\setlength{\tabcolsep}{1.7mm}
\centering
\begin{subtable}[t]{1.0\textwidth}
\begin{tabular}{cc|cc}
\toprule[1.2pt]
$\lambda_1$   & $\lambda_2$     & \textbf{OA}                & \textbf{mIoU}                            \\ \midrule
0.005 & 0.015 & 83.5  & 75.0  \\
\rowcolor{gray!15}
\textbf{0.01} & \textbf{0.015}  & \textbf{84.1} & \textbf{75.6} \\ 
0.01    & 0.010  & 83.7  & 75.3   \\
0.01  & 0.02 &  83.9  & 74.9     \\
0.015  & 0.015 &  84.0  &  75.2    \\
  \bottomrule[1.2pt]
\end{tabular}
\vspace{2pt}
\caption{Loss coefficients $\lambda$.}
\label{tab:loss}
\end{subtable}
\end{subfigure}
\vspace{-8pt}
\caption{\textbf{Effect of the filtering thresholds and loss coefficients.} }
\vspace{-2pt}
\end{table}

\begin{table}[t]
\small
\centering
\begin{subfigure}{0.235\textwidth}
\renewcommand{\arraystretch}{1.03}
\setlength{\tabcolsep}{1.0mm}
\centering
\begin{subtable}[t]{1.0\textwidth}
\begin{tabular}{c|c>{\columncolor{gray!15}}ccc}
\toprule[1.2pt]
\textbf{}  & 2000 &\textbf{4000} & 5000 & 6000 \\ \midrule
\textbf{OA}    &   82.0    & \textbf{84.1} & 83.4   &   83.0     \\
\textbf{mIoU}   & 72.1      & \textbf{75.6} & 75.2     & 74.9     \\ \bottomrule[1.2pt]
\end{tabular}
\vspace{2pt}
\caption{Warm up stages.}
\label{tab:warm}
\end{subtable}
\end{subfigure}
\begin{subfigure}{0.22\textwidth}
\renewcommand{\arraystretch}{1.03}
\setlength{\tabcolsep}{1.2mm}
\centering
\begin{subtable}[t]{1.0\textwidth}
\hspace{2em}
\begin{tabular}{c|c>{\columncolor{gray!15}}c}
\toprule[1.2pt]
               & w/o Neg    & Neg                      \\ \midrule
\textbf{OA} & 83.3  & \textbf{84.1} \\
\textbf{mIoU} & 74.6 & \textbf{75.6} \\ 
  \bottomrule[1.2pt]
\end{tabular}
\vspace{2pt}
\caption{Negative set.}
\label{tab:neg}
\end{subtable}
\end{subfigure}
\vspace{-8pt}
\caption{\textbf{Effect of the warm-up stages and negative prototype set.} w/o refers to without. }
\vspace{-2pt}
\end{table}

\subsection{More Ablation Studies}
We provide more ablation experiments in this section, and all results are reported on the PASTIS \textit{train} set. 

\noindent\textbf{Comprehensive ablation results on proposed modules.} In \cref{tab:ablation}, we present additional ablation results of our proposed modules. It can be observed that our proposed modules synergize effectively, as mentioned in the main paper. 
$\mathcal{L}_\text{cbl}$ regularizes the embedding space, facilitating the global perception of the space-time clues to crop regions. Simultaneously, $\mathcal{L}_\text{tap}$ mitigates anomalous semantics while indirectly reinforcing the stability of spatial clustering process.

\noindent\textbf{Effect of filtering thresholds $\mu$ and loss coefficients $\lambda_i$.} In Sec. \textcolor[RGB]{42,120,189}{3.2} of the main paper, we employ two thresholds $(\mu_l,\mu_h)$ to filter out the most class-relative regions, both positively and negatively, as follows:
\begin{equation}
\hat{\mathcal{M}}=\begin{cases}0,&\mathrm{if} \ \mathcal{M}\leq\mu_{l},
\\1,&\mathrm{if} \ \mathcal{M}\geq\mu_{h},
\\\mathrm{ignore},&\mathrm{otherwise}. \end{cases}
\end{equation}
\cref{tab:score} shows the performance variations under different filtering thresholds. As we can see, an excessively stringent threshold may impede the ability to capture the patterns of crops, whereas a lenient threshold may introduce undesired noise to the prototypes.
In addition, we report the impact of different loss coefficients on accuracy, the results are presented in \cref{tab:loss}.

\begin{figure*}[t]
\centering
\includegraphics[width=0.8\linewidth]{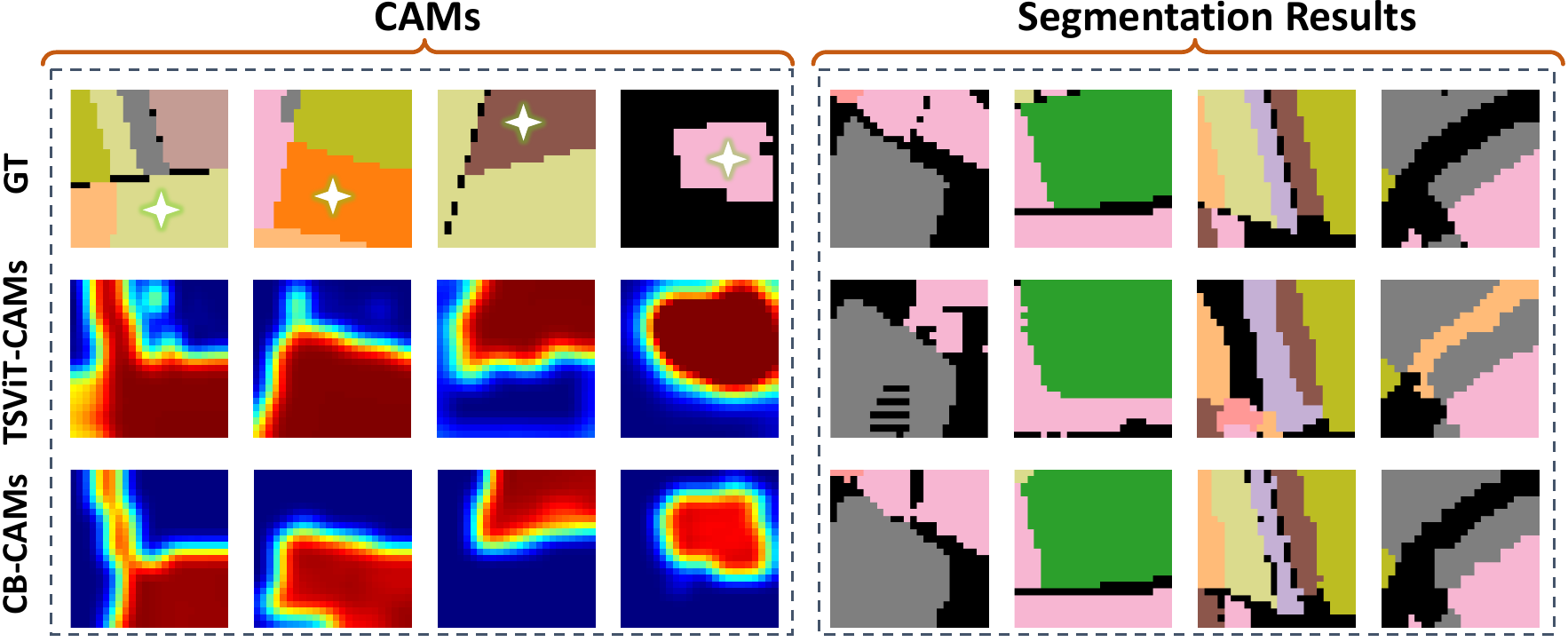}
\caption{\textbf{Qualitative results between baseline TSViT-CAM and CB-CAMs derived by \textit{Exact} on Germany dataset.} \textbf{Left}: CAMs comparisons. \textbf{Right}: Semantic segmentation comparison results. 
The stars represent the corresponding activation category.
}
\label{fig:cams}
\end{figure*}

\begin{figure}[t]
\centering
\includegraphics[width=1.0\linewidth]{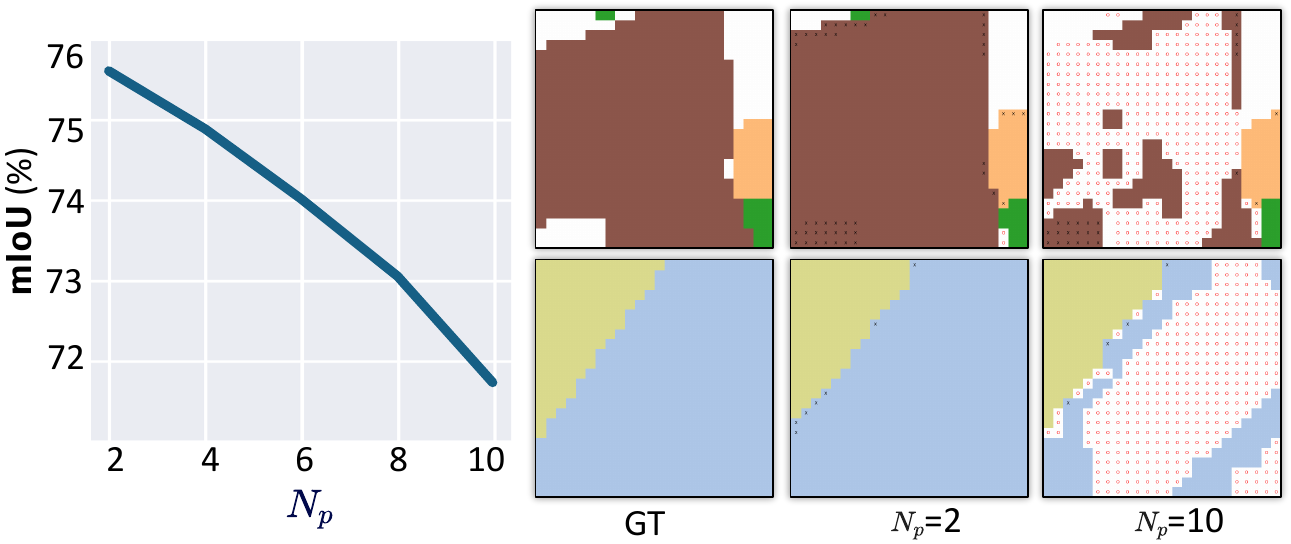}
\caption{\textbf{Quantitative and qualitative results of pseudo labels with different $N_p$.} The red dot ${\color{Red}\circ }$ and black cross $\times$ in qualitative results denote the false negative and the false positive activations, respectively.} 
\label{fig:proto}
\end{figure}

\noindent\textbf{Effect of the warm up stage.}
The prototype learning relies on the raw CAM's accurate perception of parcel objects.
Since the network lacks the capability to perceive parcel objects at the early training stages, prematurely introducing prototype learning and feature space shaping may result in gradient explosion. 
\cref{tab:warm} shows the impact on the warm up stages for the model performance. We can see that starting the prototype learning and clue-based contrastive learning at 4000 iterations can achieve the best performance. 
While an excessively prolonged warm-up stage may cause the model to memorize noise perturbations, thereby hindering the shaping of the feature space.

\noindent\textbf{Effect of the negative set.} 
In the main paper, we introduce a positive prototype set and a negative prototype set to capture class-relevant positive and negative patterns, respectively. We present additional quantitative results in \cref{tab:neg} to demonstrate the effectiveness of the negative set. The results indicate that the negative class-relevant semantics can complement with positive patterns, thereby assisting the model in eliminating erroneous crop regions.

\begin{figure}[t]
\centering
\includegraphics[width=1.0\linewidth]{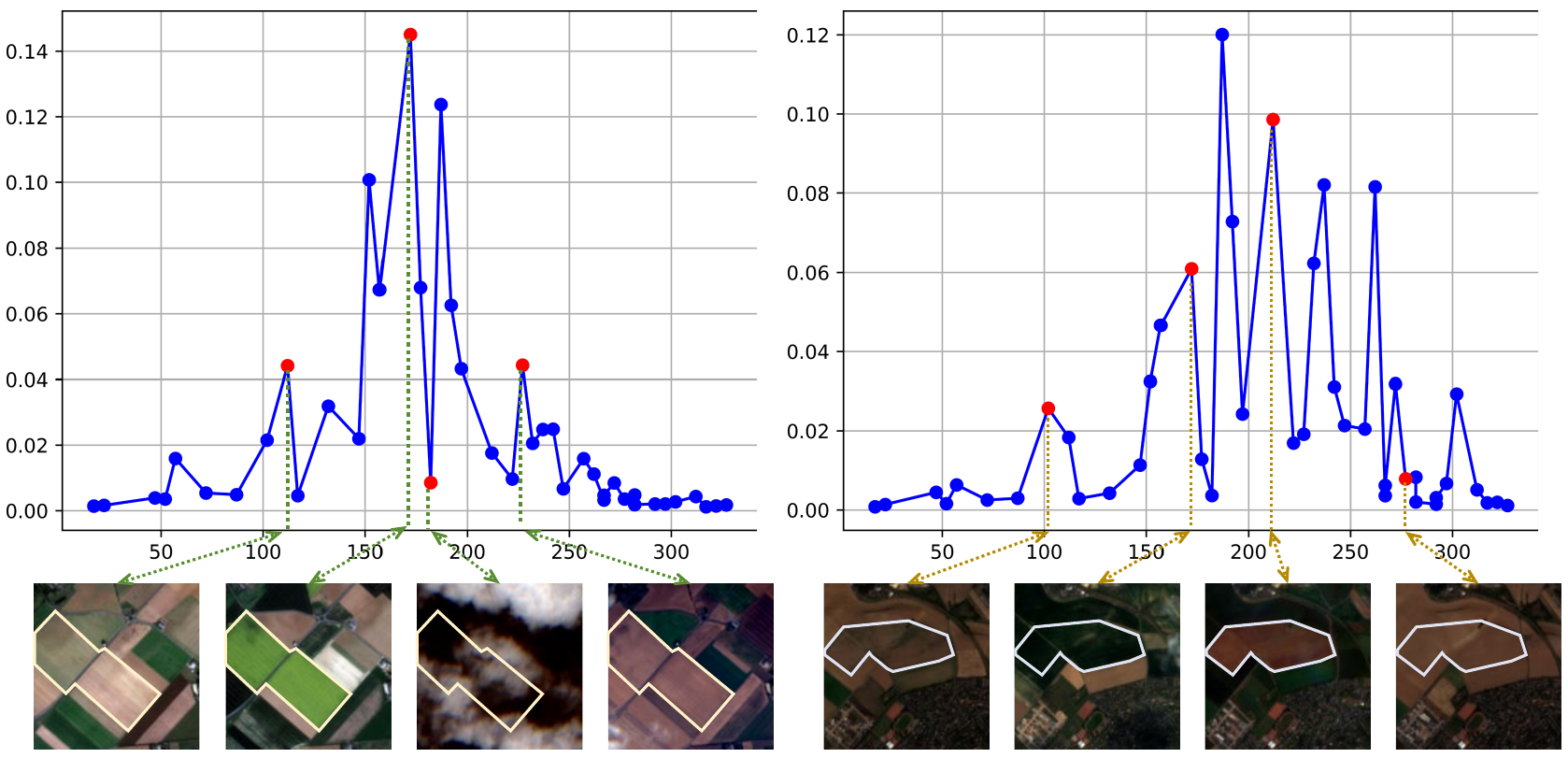}
\caption{\textbf{Single satellite image extracted by temporal-to-class attention.} The line chart represents the temporal-to-class attention, and the red dots correspond to the satellite images shown below.}
\label{fig:temporal}
\end{figure}

\noindent\textbf{Adverse impact of increasing the number of prototypes.} As discussed in the main paper, an excessive number of prototypes may impair the model's ability to perceive the global unified semantics of the crop parcel. In \cref{fig:proto}, we present both quantitative and qualitative experimental results to illustrate the impact of increasing the number of prototypes. As we can see, the prototypes' ability to perceive crop parcels declines sharply as $N_p$ increases to 10 (75.6\% vs. 71.8\% mIoU). This is mainly due to the large $N_p$ compels the prototypes to capture local discriminative patterns, thereby resulting in a severe under-activation issue.

\subsection{Additional Qualitative Results}
\noindent\textbf{Low-level mapping of the temporal-to-class attention.} In order to intuitively demonstrate the effect of  temporal-to-class attention on the perception of temporal sequences, we list the satellite images under different attention scores. As shown in \cref{fig:temporal}, temporal clips with high attention scores contain pivotal information for crop recognition, whereas those low scores are associated with anomalous temporal periods (\eg, cloud cover, barren land). Therefore, explicitly emphasizing the contributions of different temporal clips to crop recognition can mitigate the confusion arising from anomalous semantics.

\noindent\textbf{Visual comparison of CAMs and segmentation results.} We additionally provide visual comparison between the TSViT-CAMs (baseline) and the proposed CB-CAMs on Germany dataset, as shown in \cref{fig:cams}. The first four columns show the visualization of the CAMs. One can observe that the CB-CAMs generated by \textit{Exact} are capable of accurately delineating the crop regions. Therefore, the semantic segmentation model tends to show more powerful perceptual capability under \textit{Exact}-generated pseudo labels' supervision.

\begin{figure*}[t]
\centering
\includegraphics[width=1.0\linewidth]{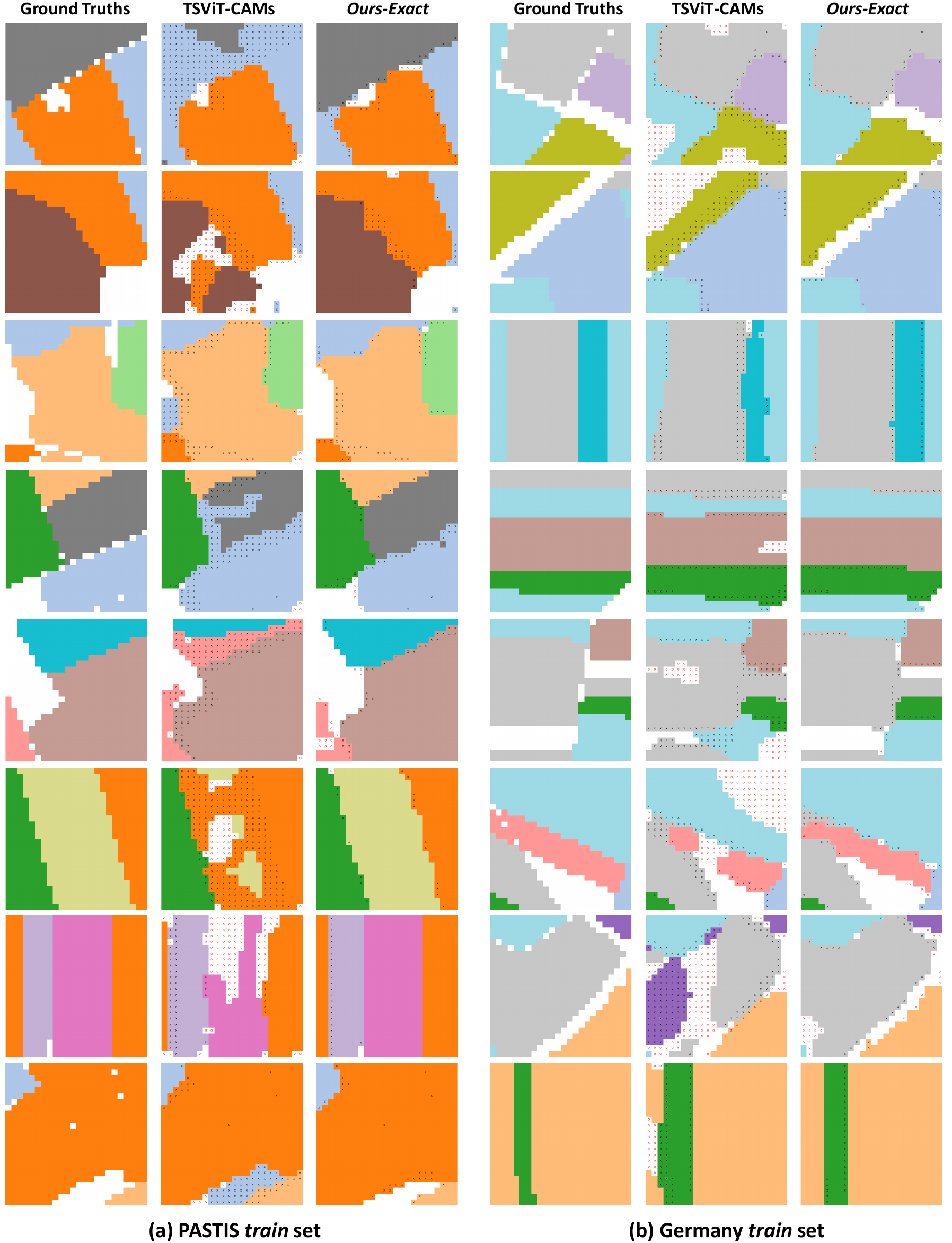}
\caption{ \textbf{Qualitative comparison of pseudo labels among baseline TSViT-CAMs and ours \textit{Exact} on PASTIS and Germany \textit{train} set.} The red dot ${\color{Red}\circ }$ and black cross $\times$ in qualitative results denote the false negative and the false positive activations, respectively.}
\label{fig:pseudo}
\end{figure*}

\noindent\textbf{Visual comparison of pseudo labels.} 
In \cref{fig:pseudo}, we show the pseudo labels derived by TSViT-CAMs and CB-CAMs on both PASTIS and Germany \textit{train} set. Consistent with the main paper, we observe that \textit{Exact} remarkably addresses both under- and over-activation issues in the baseline, thereby providing more reliable supervision for SITS semantic segmentation network.

